\documentclass[twoside,11pt]{article}

\usepackage{blindtext}

\usepackage[preprint]{jmlr2e}
\usepackage[utf8]{inputenc}
\usepackage{amsmath,amsfonts}
\usepackage{algorithm}
\usepackage{algorithmicx}
\usepackage{algpseudocode}
\usepackage{array}
\usepackage[caption=false,font=normalsize,labelfont=sf,textfont=sf]{subfig}
\usepackage{textcomp}
\usepackage{stfloats}
\usepackage{url}
\usepackage{verbatim}
\usepackage{graphicx}
\usepackage{comment}
\usepackage{booktabs}
\usepackage{multirow}
\usepackage{tabularx}
\usepackage{threeparttable}
\usepackage{pdflscape} 
\usepackage{enumitem}
\usepackage{mathtools}
\usepackage{pdflscape} 
\usepackage{rotating}  

\newcommand{\meanpm}[2]{%
  \if\relax\detokenize{#1}\relax
  \else
    #1 \text{\tiny{(#2)}}%
  \fi
}

\newcolumntype{Y}{>{\centering\arraybackslash}X} 
\newcolumntype{L}{>{\raggedright\arraybackslash}X} 

\DeclareMathOperator*{\argmin}{argmin}

\usepackage{lastpage}
\jmlrheading{23}{2026}{1-\pageref{LastPage}}{3/26}{tbd}{21-0000}{Daniel Schweizer, Peter Kuhn, Jayant Sharma, Shivali Dubey, Malte von Ramin and Christoph Brockt-Ha{\ss}auer}

\ShortHeadings{Distribution-Aware Conformal Prediction}{Schweizer, Kuhn, Sharma, Dubey, von Ramin and Brockt-Ha{\ss}auer}
\firstpageno{1}

\begin{document}

\title{Distribution-Aware Conformal Prediction: A Framework for Generating Efficient Prediction Intervals for Time Series}

\author{\name Daniel Schweizer$^\dagger$  \email daniel.schweizer@emi.fraunhofer.de \\
        \name Peter Kuhn$^\dagger$  \email peter.kuhn@emi.fraunhofer.de\\ 
        \name Jayant Sharma\\ 
        \name Shivali Dubey\\ 
        \name Malte von Ramin\\ 
        \name Christoph Brockt-Ha{\ss}auer\\
       \addr Fraunhofer Institute for Highspeed Dynamics, Ernst-Mach-Institut, EMI\\
       Freiburg, Germany\\
       $^\dagger$\small{These authors contributed equally.}}

\editor{To be assigned.}

\maketitle

\begin{abstract}
We present Distribution-aware Conformal Prediction (DCP), a unified framework integrating probabilistic predictors like Monte Carlo dropout, deep ensembles, and quantile regression with score--agnostic conformal calibration to produce valid and efficient prediction intervals. Leveraging a numerical inversion approach to construct interval bounds, DCP accommodates arbitrary combinations of distribution generating predictors and nonconformity scores. Benchmark analysis on synthetic and real--world time series data demonstrate DCP's ability to adaptively calibrate prediction intervals under varying uncertainty regimes. Crucially, DCP’s modular design facilitates plug--and--play experimentation with different predictor--score pairings, quantitatively supported by a newly introduced modified Winkler score that balances validity and efficiency by explicitly penalizing undercoverage. While DCP generalizes and extends existing approaches like Conformalized Quantile Regression and Conformalized Monte Carlo, its modular design allows further extensions, setting a foundation for advancing uncertainty quantification in dynamic environments and high-risk applications.
\end{abstract}


\section{Introduction}

Standard neural networks yield point forecasts without any intrinsic measure of predictive uncertainty, a limitation that becomes critical in high-risk applications such as energy, traffic, and finance \citep{nowotarski2018, bazionis2021, Li2017DiffusionCR, BLASCO2024127339}. In these domains, misestimated uncertainty can trigger blackouts, congestion, or large financial loss. Recognizing these stakes, the EU Artificial Intelligence Act therefore classifies such systems as high-risk and mandates that providers disclose accuracy, expected error rates, and performance limitations~\citep{EUAIAct2025}, echoing the long-standing view that forecasts must be accompanied by statistically reliable, well-calibrated uncertainty estimates. Poorly calibrated predictive intervals (PI) can be as misleading to decision-makers as no uncertainty information at all. In this paper we present a framework that jointly produces predictive distributions and calibrates them rigorously, aligning technical soundness with emerging regulatory requirements.

To quantify predictive uncertainty, a variety of techniques for augmenting standard neural networks have emerged. For example, Monte Carlo dropout (MCD)~\citep{gal2016} resamples a random dropout pattern that zeros a subset of activations at each forward pass, while deep ensembles~\citep{lakshminarayanan2017} train multiple networks from different initializations. These methods yield a predictive distribution whose variance captures mainly \textit{epistemic uncertainty}, i.e., ignorance about the data-generating process~\citep{huellermeier_2021}. Such estimates may aid Out-of-Distribution (OOD) detection, where large predictive variance flags novel inputs. Yet they lack formal coverage guarantees and therefore require calibration.

Conformal prediction (CP) is a model-agnostic, well-studied statistical framework that wraps any base predictor to produce distribution-free prediction intervals (PI), while guaranteeing finite--sample marginal coverage under exchangeability~\citep{vovk2022}. Applied to a deterministic point predictor, CP yields valid but often conservative PIs, because their width is governed by a single global calibration threshold, with no input--specific adaptation~\citep{Bethell_Gerasimou_Calinescu_2024, fan2024}. In contrast, probabilistic predictors provide input--dependent uncertainty arising from data noise and, when modeled appropriately, from limited knowledge. Embedding such predictors in a conformal wrapper with nonconformity scores that leverage the predictive distribution can produce adaptive and \textit{efficient} intervals. Several studies have paired CP with specific methods like MCD or deep ensembles~\citep{mediyev2025, Bethell_Gerasimou_Calinescu_2024,romano2019,angelopoulos2023, Chen2021,Jensen2024}. However, most of these hybrid approaches are ad hoc; they fix a one nonconformity score for a particular predictor and application domain, and without a unified framework to compare choices across predictors, scores, and data regimes, they remain narrow in scope. Our objective is to provide a systematic framework that pairs distribution-generating predictors (DGP) with conformal calibration and guides practitioners in selecting predictor-score combination that maximize PI efficiency.

We introduce \emph{Distribution--aware Conformal Prediction} (DCP), a general framework that fuses probabilistic predictors with conformal calibration, yielding PIs that are both valid and sharp---hence efficient. DCP treats any DGP such as MCD, deep ensembles or quantile regression networks as a black box and incorporates their uncertainty estimates via model--agnostic nonconformity scores. Interval bounds are obtained via a bracketing and bisection root-finding algorithm as an alternative to analytical inversion, making the procedure plug--and--play for arbitrary, possibly asymmetric or nonmonotone, scores, and reproducing closed‑form cases up to numerical tolerance. To showcase the algorithm and the benefits of different predictor--score combinations, we study three generic score families: (i) error-based scores such as the absolute residual, (ii) interval-violation scores that quantify where the target lies with respect to the inner bands of a predictive distribution, and (iii) density-based scores implemented via a K-Nearest Neighbor local-density surrogate in the predictive output space. We benchmark these DGP-score combinations using three sequence models on six real-world time series data sets spanning energy, mobility, finance, and mixed-domain settings. Furthermore, we use synthetic data to isolate the effects of epistemic drift and heteroscedastic noise. To quantify the trade--off between coverage and sharpness we propose a modified version of the Winkler score~\citep{winkler1972}, an efficiency metric that penalises under-coverage. The main contributions of this paper are:

\begin{enumerate}

\item{A unified framework (DCP) that combines DGPs with arbitrary nonconformity scores under conformal calibration to produce efficient PIs for time series.}

\item{A score--agnostic numerical inversion method to construct interval bounds, as an alternative to analytical inversion, avoiding score--specific algebra.}

\item{A modified mean Winkler (MMW) metric that combines width and miss distance with a tolerance--adjusted undercoverage penalty to assess interval efficiency.}

\item{Extensive experiments on synthetic and six real-world data sets, providing practical guidance for predictor--score selection while highlighting aleatoric and epistemic uncertainty regimes.}

\end{enumerate}

Beyond these individual components, DCP also serves as a unifying lens on existing conformalized uncertainty methods. By choosing appropriate combinations of distribution--generating predictors and nonconformity scores, DCP recovers the functional forms of several established approaches such as Conformalized Quantile Regression (CQR)~\citep{romano2019}, Conformalized Monte Carlo (CMC)~\citep{mediyev2025}, and variance--scaled Monte Carlo dropout methods in the spirit of MC-CP~\citep{Bethell_Gerasimou_Calinescu_2024}. At the same time, the framework extends these methods by enabling additional, previously ad hoc combinations (e.g., density-based scores on ensemble predictors) within a single, score--agnostic calibration pipeline.

We focus on time series forecasting, which poses particular challenges for CP frameworks, because temporal dependence often violates data exchangeability. To address this, we employ an online sliding--window variant of split conformal prediction that tracks drift and aim for approximate marginal coverage (see Section~\ref{sec:ExperimentalSetup}). We evaluate DCP’s robustness empirically and analyze its behavior in Section~\ref{sec:ResultsAndDiscussion}. DCP is task agnostic, which means that any predictor that outputs either point estimates or full predictive distributions can be wrapped to obtain valid prediction sets. A complete list of symbols and notation is provided in the Supplementary Material D.

\section{Related Work} 
\label{sec:related_work}
\noindent Distribution-aware Conformal Prediction (DCP) combines Conformal Prediction (CP) with uncertainty quantification techniques that output a predictive distribution. This section is therefore organized into three parts: (i) DGPs for uncertainty estimation, (ii) a concise recap of conventional CP, and (iii) existing hybrid approaches that tie the two together.

\subsection{Distribution-Generating Predictors}
\noindent Predictive uncertainty is usually decomposed into an \emph{aleatoric} component, due to irreducible randomness inherent in the data, and an \emph{epistemic} component, which stems from limited knowledge about the data--generating process. Both are reflected in the spread of the predictive distribution. To span these regimes and evaluate DCP under distinct modeling assumptions, we adopt four representative, widely used DGPs: (i) Quantile Regression~(QR)~\citep{koenker1978}, which learns conditional quantiles directly and therefore targets aleatoric variability; (ii) Monte Carlo dropout~(MCD)~\citep{gal2016}, which quantifies epistemic uncertainty by resampling dropout masks at inference; (iii) Bootstrap ensembles~(BE)~\citep{hastie2009}, which use data resampling to capture epistemic uncertainty; and (iv) Deep ensembles (DE)~\citep{lakshminarayanan2017}, which train multiple networks with different random initializations and likewise measure predominantly epistemic variability.

The literature contains many additional DGPs such as batch-normalization uncertainty~\citep{teye2018}, DropConnect~\citep{dropconnect}, snapshot ensembles~\citep{HuangLP0HW17}, additive Gaussian noise~\citep{yuan2025}, Gaussian neural networks~\citep{kuhn2025gaussian}, any of which could, in principle, serve as the backbone in DCP. Although, their practical utility will depend on the match between uncertainty type and data regime as well as the predictors architecture, and computational budget. We focus on QR, MCD, BE and DE because they span distinct sources of uncertainty and are widely adopted in practice.

\subsubsection{Quantile Regression}
As an extension to linear regression it learns conditional quantiles (e.g., the 5th percentiles) by minimizing pinball loss at selected levels of \(\theta\)~\citep{taylor2000, takeuch2006, rodrigues2020,koenker1978}. Each estimated quantile gives a value below which the target falls with probability \(1-\theta\). Combining two quantiles yields PIs that adapt to heteroscedastic data and capture aleatoric uncertainty~\citep{melba:2022:008:akrami}. These intervals, however, lack finite-sample coverage guarantees~\citep{Sesia2020}.

\subsubsection{Monte Carlo Dropout}
Dropout was introduced as a regularizer that randomly sets a fraction of activations to zero during training~\citep{srivastava14a}. Gal and Ghahramani~\citep{gal2016} later showed that keeping dropout active at inference and performing multiple stochastic forward passes known as Monte Carlo dropout (MCD), is equivalent to variational inference in a Bayesian neural network, thus effectively drawing samples from the posterior predictive distribution with little additional computational overhead. For such sampling-based methods the sample mean serves as the point forecast and the sample variance quantifies epistemic uncertainty. 

\subsubsection{Ensembles}
Bootstrap aggregating (bagging) ensembles train multiple identical instances of a base predictor on bootstrap resamples of the data~\citep{kim2020,hastie2009}. While some data points are drawn repeatedly, others will be left out, constituting the out-of-bag set. Deep Ensembles, on the other hand, train the same architecture independently, with different random initialization of weights (and, optionally, other minor variations) to induce diversity~\citep{lakshminarayanan2017}. In both cases the resulting collection of predictions is the predictive distribution.

Empirically, ensembles tend to provide better calibrated uncertainty than MCD~\citep{lakshminarayanan2017} and may be used to detect outliers~\citep{Ovadia2019}, albeit at higher computational and memory cost.

\subsection{Conformal Prediction}
\noindent While DGPs yield predictive distributions, intervals formed from their nominal quantiles (e.g., central or highest-density regions) often lack correct coverage, due to poor calibration. Conformal prediction (CP) addresses this by providing a model-agnostic, distribution-free framework that converts any predictor into PIs with a rigorous marginal coverage guarantee~\citep{vovk2022}.

CP assumes that the calibration and test observations are \emph{exchangeable}, an assumption slightly weaker than independent and identically distributed (i.i.d.). It requires a real-valued \emph{nonconformity score} \(s(y,\hat{y}(x))\), which quantifies how unusual a prediction \(\hat{y}(x)\) is relative to the true value \(y\).

The original "full" CP method refits the model for every candidate label and retains those labels whose scores fall below a prescribed quantile. While conceptually simple, full CP is computationally infeasible for modern neural networks. Instead, split CP~\citep{Lei03072018} is generally being used as an alternative that trains the model once, evaluates a nonconformity score on a holdout \emph{calibration} set, and uses the empirical \((1-\alpha)\)-quantile \(\hat q\) of the score results to build intervals for a given nominal miscoverage rate $\alpha \in [0,1]$). For regression tasks, the canonical nonconformity score is the absolute residual 
\begin{equation}
  s(y,\mathbf{\hat y }(x))=\lvert y-\hat\mu(x)\rvert ,  
\label{eq:abs_res}
\end{equation}
\noindent where \(y\) is the ground truth, \(\mathbf{\hat y } (x)\) is the predictive distribution and \(\hat\mu(x)\) is the point prediction (or mean value of \(\mathbf{\hat y } (x)\)) produced by the model for input \(x\). This score yields the symmetric interval \([\hat\mu(x)-\hat q,\;\hat\mu(x)+\hat q]\).

Let \(\Bar{\mathcal{C}}\) denote the split conformal prediction interval obtained with the calibration scores. Then, regardless of the specific variant, we have
\begin{equation} 
\Pr\left(y \in \Bar{\mathcal{C}}\left( \mathbf{\hat y} (x)\right)\right) \geq 1 - \alpha,\ \forall (x,y) \in D^{\text{t}} .
\end{equation}
\noindent The guarantee is \emph{marginal}: it holds on average over the test set \(D^t\). A comprehensive tutorial appears in \citep{angelopoulos2023}, and formal proofs can be found in \citep{Lei03072018}.

Popular extensions to CP address common practical challenges such as limited effective sample size, non‑stationary data, and robustness. Resampling--based batch methods, including jackknife+ and cross--validation+~\citep{barber2020}, recycle training data to tighten intervals while maintaining coverage, whereas adaptive approaches, including online and sliding window conformal methods, update the calibration set to handle distribution shift~\citep{gibbs2021}. Beyond exchangeability, weighted quantiles and randomization have been proposed to improve robustness to distributional drift~\citep{barber2023}. Messoudi et al.~\citep{messoudi2020} target robustness to noisy and adversarial inputs, demonstrating that conformal prediction can enhance reliability in critical decision making contexts. Complementary to these extensions, Alijani and Najjaran propose WQLCP, a vision--focused CP variant that leverages VAE reconstruction losses to re--weight calibration quantiles and scale scores, recovering coverage under distribution shifts~\citep{alijani2025}.

The flexibility of choosing any real--valued nonconformity score allows CP to inherit information from distribution--generating predictors. Section~\ref{sec:nonconformityscores} details the specific scores we employ in our framework.

\subsection{Conformalized Distribution Generating Predictors}
\noindent Several studies have already fused distribution-generating predictors with CP. Each of their methods exploits different aspects of the predictive distribution and therefore makes a specific trade-off between computational cost, distributional assumptions and interval sharpness.

A first family augments Monte Carlo dropout. MC-CP~\citep{Bethell_Gerasimou_Calinescu_2024} replaces the fixed split-conformal width by a locally adaptive factor proportional to the the predictive variance. Implementation is trivial and the intervals retain finite-sample coverage under exchangeability, but the method implicitly assumes approximate symmetry of the predictive errors around the mean using a residual based score and thus discards potential shape information in the sampled distribution on skewness or multimodality. Conformalized Monte-Carlo (CMC)~\citep{mediyev2025} moves beyond symmetry by extracting a highest-density interval (HDI) from the MCD samples and calibrating a single global slack term (Section~\ref{sec:intervalscore}); the same HDI score can, in principle, be computed from any DGP, a flexibility we exploit in Section \ref{sec:ExperimentalSetup}. CMC works well for asymmetric predictive distributions, yet its efficiency naturally hinges on how faithfully the DGP captures the true spread.

Conformalized Quantile Regression (CQR)~\citep{romano2019,angelopoulos2023, Sesia2020} fits a model using pinball loss to get lower and upper conditional quantiles, then conformalizes them on a held-out calibration set by adding a single global slack equal to the empirical quantile of the respective interval--violation score. This yields locally adaptive intervals for heteroscedastic aleatoric noise when the quantiles are informative. Because CQR does not model epistemic uncertainty, coverage can degrade under distribution shift.

Ensemble Batch Prediction Intervals (EnbPI)~\citep{Chen2021} use leave-one-out forecasts from a bagged model and calibrates them in a sliding window, yielding valid intervals under mild mixing conditions without a separate calibration split. Because  only the interval width is updated online (i.e., recalibrated with new residuals) while its center remains anchored to an ensemble predictor frozen after training, EnbPI adjusts only slowly to abrupt local changes in mean or variance. Ensemble Conformalized Quantile Regression (EnCQR)~\citep{Jensen2024} remedies this by replacing point forecasts  with ensemble-based lower and upper quantile functions and keeps the windowed recalibration. The resulting intervals track both heteroscedasticity and concept drift, remaining sharper than EnbPI while preserving coverage.

Probabilistic Conformal Prediction (PCP) and Generative Conformal Prediction with Vectorized Nonconformity Scores (GCP-VCS) extend CP to generic generative models~\citep{zheng2025}, by constructing separate intervals around each sampled point in the predictive distribution, which is effective for multimodal targets. GCP-VCS uses average nearest neighbor distance internally, however, its nonconformity score is the residual.

These studies demonstrate that individual DGPs can indeed be conformalized, but each solution is tailored to either a specific predictor--score combination, or sometimes even particular data regime. None provide a unified, score-agnostic framework that integrates arbitrary combinations of distribution--generating predictors and nonconformity scores, and enables principled comparison with an efficiency metric. Addressing this gap is precisely the aim of our Distribution--aware Conformal Prediction (DCP) framework.

MAPIE~\citep{cordier2023} exemplifies efforts to create flexible and accessible frameworks for conformal prediction, offering seamless integration with scikit-learn models for common machine learning tasks. However, it does not explicitly support direct integration with distribution-generating predictors or means for systematic comparison.

From the perspective of DCP, many of these methods correspond to particular choices of DGP and nonconformity score under split conformal calibration. For example, CQR arises when the DGP is a quantile--regression model and the score is an interval--violation measure based on conditional quantiles. CMC is recovered by using a Monte Carlo dropout predictor with a highest-density interval (HDI) extracted from the draws and an associated interval--violation score. MC-CP–style methods correspond to Monte Carlo dropout paired with a variance--scaled residual (Z-score). DCP generalizes these constructions by allowing arbitrary DGP--score pairings, including shape--sensitive density--based scores such as KNN surrogates on ensemble predictors, within a single numerical inversion back end.

\section{Distribution-aware Conformal Prediction} 
\label{sec:DCP}
\begin{figure*}[t]
\centering
\includegraphics[width=\textwidth]{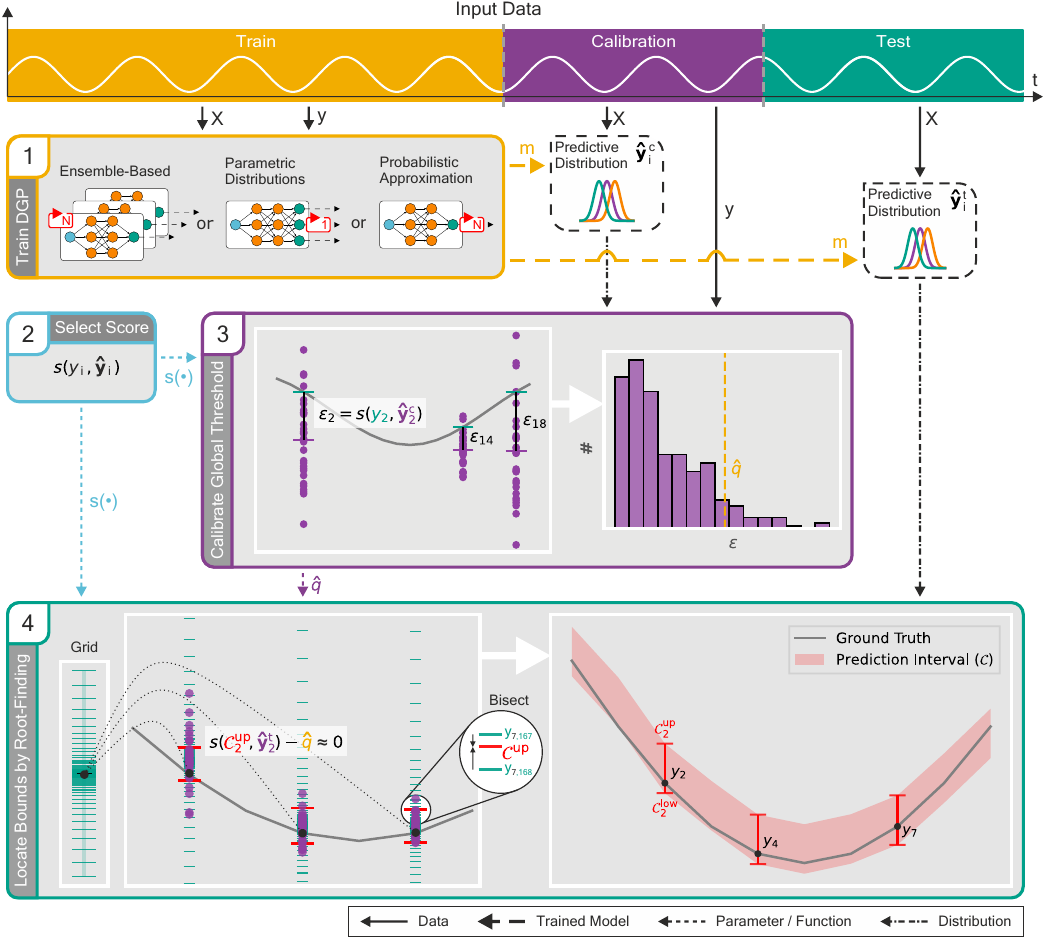}
\caption{A schematic overview of the DCP framework: (1) Start by training a DGP to generate predictive distributions on the calibration and test set. (2) Select a nonconformity score that evaluates how uncharacteristic a ground truth is for a predictive distribution. (3) Calibrate a global threshold $\hat{q}$ on the calibration set and (4) locate the prediction intervals for the test set by root-finding, which determines the values along a grid for which the nonconformity is approximately equal to the threshold.}
\label{fig:framework}
\end{figure*}

\noindent Our framework extends split CP by leveraging the full shape of the DGP's predictive distribution. Instead of adjusting a single point prediction with a constant, symmetric width, DCP calibrates a \emph{distribution-aware} nonconformity score, yielding adaptive PIs that can be asymmetric, heteroscedastic or, in principle, also multimodal, as implied by the underlying predictive distribution. The framework comprises four conceptual steps (illustrated in Fig.~\ref{fig:framework} and formalized in Algorithm~\ref{alg:DCP}):

\begin{enumerate}[label=(\roman*)]
\item \textbf{Train a distribution-generating predictor:}
Fit any model that outputs a predictive distribution $\hat P_{x}$ for every input \(x\),
\begin{equation}
      \hat P_{x}(y)\;=\;p_\theta(y\mid x),
\end{equation}
where $p_\theta(y\mid x)$ is the conditional distribution over \(y\) for a given \(x\) produced by the model with parameters \(\theta\), capturing the uncertainty of the prediction. Regardless of whether $\hat P_{x}$ is available in closed form (e.g., Gaussian process) or only implicitly (e.g., MCD, ensembles or QR), we draw a fixed number \(M\) of samples and collect them in the \emph{draw vector}
\begin{equation}
      \mathbf{\hat{ y}}(x) 
      = \bigl[\left(\hat y(x)\right)_{j}\bigr]_{j=1}^{M}\;\subseteq\;\hat P_{x}.
\end{equation}

\item \textbf{Select a distribution-aware score:}
Choose a real-valued nonconformity score \(s(y,\mathbf{\hat{ y}}(x))\) that measures how atypical the candidate outcome \(y\) (ground truth during calibration, hypothetical value during inference) is with respect to the predictive distribution. The absolute residual in ~\eqref{eq:abs_res} recovers classical split CP, whereas more sophisticated, distribution-aware scores may exploit the full shape information provided by the DGP in order to adapt to local scale, skewness, or multimodality. Any real-valued score can be paired with any DGP from step (i); the specific families we evaluate are introduced in Section~\ref{sec:nonconformityscores}.

\item \textbf{Calibrate a global threshold:}  
Hold out a calibration set \(D^{c}=\{(x^c_i,y^c_i)\}_{i=1}^{N^c}\). From here on we use the upper index on inputs, ground truth and draw vector to specify the part of the data set they belong to, i.e., \((x_i^d, y_i^d) \in D^d\) for \(d \in \{c,t\}\), specifying the calibration and test set, and \(\mathbf{\hat{ y}}_i^d = \mathbf{\hat{ y}}(x_i^d)\).

For each sample \((x^c_i,y^c_i)\) generate its predictive distribution \(\mathbf{\hat{ y}}_i^c \subseteq m(x_i)\), where \(m\) denotes the trained DGP from step (i), and compute the nonconformity score values
\begin{equation}
\varepsilon_i \;=\; s\bigl(y_i,\mathbf{\hat{ y}}_i^c\bigr).
\end{equation}
Let \(\hat q\) be the empirical \((1-\alpha)\)-quantile of the score collection \(\{\varepsilon_i\}_{i=1}^{N^c}\); equivalently, \(\hat q\) is the \(\lceil (N+1)(1-\alpha) \rceil\)-th smallest score. By construction at least \(100(1-\alpha)\%\) of the calibration points satisfy \(\varepsilon_i \le \hat q\).

\item \textbf{Locate the intervals by root-finding:}
For a new test input \(x_i^t\), define the predictive set
\begin{equation}
    \mathcal{C}_i
    =\bigl\{y\in \mathbb{R}\;:\; s\bigl(y,\mathbf{\hat{ y}}_i^t\bigr)\le\hat q \bigr\},
\end{equation}
\noindent and numerically compute its interval boundary points by solving
\begin{equation}
\label{eq:root}
f_i(y)=s\bigl(y,\mathbf{\hat{ y}}_i^t\bigr)-\hat q=0.
\end{equation}
Anchor the search at the predictive median \(\tilde y_i\), and build an ordered, symmetric geometric grid \(\tilde y_i + \{0,\pm h_0,\pm h_0\gamma,\ldots\}\) up to a capped depth (using empirically chosen default parameters \((h_0,\gamma,\mathrm{tol},\mathrm{depth})\) in Table~\ref{tab:root_params}, which are not claimed to be optimal). Evaluate \(f_i\) on this grid and detect sign changes. Then take the outermost sign changes (leftmost and rightmost adjacent pairs) as the brackets and refine each bracket by bisection to an absolute tolerance, yielding the interval endpoints \(\mathcal{C}^{\text{low}}_i\) and \(\mathcal{C}^{\text{up}}_i\), such that
\begin{equation}
\label{eq:bounds}
    \mathcal{C}_i \approx \bigl[\mathcal{C}^{\text{low}}_i, \mathcal{C}^{\text{up}}_i \bigr].
\end{equation}
This construction is agnostic to the sign of \(\hat q\), allowing intervals to expand or shrink accordingly.
\end{enumerate}

\subsection{Analytic Bounds and Generic Root-Finding}
\noindent Under split conformal prediction with a nonconformity score that is monotone in \(y\), several choices admit analytic inversion, yielding closed-form bounds for the conformal set. For the residual score based on a point predictor \(\hat{\mu}\), with scaling being applied, the conformal prediction interval is
\begin{equation}
\label{eq:analy_inverse}
\Bar{\mathcal{C}}_i = \hat{\mu}(x_i^t)\pm \hat{q}\,u(x_i^t),
\end{equation}
\noindent where \(u(x_i^t)\) (\(u(x_i^t) > 0\)) is a \(y\)-independent uncertainty scalar, such as the local standard deviation \(\sigma\), computed from the draw vector~\citep{angelopoulos2023}. Likewise, for a interval (violation) score built from inner bands, analytic inversion yields
\begin{equation}
\label{eq:int_inverse}
\Bar{\mathcal{C}}_i = [\,\mathcal{C}^{\text{low}}_i - \hat q\,u(x_i^t),\;\mathcal{C}^{\text{up}}_i + \hat q\,u(x_i^t)\,].
\end{equation}
Some distribution-aware scores are density-based, i.e., they increase as the estimated predictive density at a candidate value \(y\) decreases. Such scores can be strongly nonlinear and nonmonotone in \(y\) (e.g., under multimodal predictive distributions), which precludes a simple analytic inverse for \(s(y,\mathbf{\hat{ y}}(x)=\mathbf{\hat{ y}}_i^t)\). Examples include nearest neighbor--based density surrogates or smooth density estimates; the exact estimator is not crucial for our method. The \emph{bisect-based root-finder} handles these cases uniformly; when closed-form bounds exist it reproduces them up to numerical tolerance, and it seamlessly extends to density-based scores without requiring score-specific algebra.

\subsection{Limits and Failure Policy}
\noindent In practice, heavy--tailed predictive distributions may push the first sign change far outside the initial grid, while multimodal or step--like scores can produce closely spaced or multiple crossings (roots) that challenge the simple bracketing. The bisect back--end therefore includes an automatic retry strategy: if fewer than two sign changes are detected on the initial geometric grid, shrink the initial step \(h_0\) and increase the grid depth to widen coverage and refine resolution, then re--evaluate the score. However, if exactly one sign change persists after the retry strategy, return a degenerate interval whose endpoints coincide at the single root. If no sign change is found, place a synthetic bracket at the predictive median \(\tilde y\) and return a degenerate interval centered there. When more than two sign changes occur (nonmonotone scores, e.g., under multimodality), select the \emph{outermost} pair of crossings to construct the interval. The latter preserves marginal coverage guarantees but may be conservative in width.

\subsection{Practical Advantages}
\noindent Taken together, the bisect--based root-finder turns conformal calibration into a clean predictor--score interface with three practical advantages. First, it is \emph{score-agnostic} which means that the nonconformity function \(s(\cdot)\) is treated as a black box, so new scores can be adopted without bespoke algebra. When closed-form bounds exist (residual, scaled residual, interval-violation), the routine reproduces them up to numerical tolerance, while also accommodating negative \(\hat{q}\) (PI shrinkage). Second, it is \emph{robust}. Because bracketing relies only on sign changes, the method remains stable under heavy tails, skewness, and non-smooth scores. In nonmonotone cases it returns the outermost crossings, yielding conservative (wider) yet valid intervals. Note that returning every root segment is straightforward but left for future work. Finally, the routine is \emph{efficient}. Per test point the cost is \(O(K\log \mathrm{tol}^{-1})\) score evaluations, with a small constant \(K\) determined by the grid expansion; the implementation is fully vectorized over horizons and features, memory-efficient, and trivially batched across samples. These properties hold for both static split CP (fixed \(\hat{q}\)) and the sliding-window adaptive variant where \(\hat{q}\) is recomputed after each revealed test target. 

\begin{table}[h]
\centering
\begin{tabular}{lccl}
\toprule
Variable & Default & Typical range & Description\\
\midrule
$\mathrm{depth}$   & 100           & \(\ge 1\)                 & Number of draws per query.\\
$\mathrm{h_0}$     & \(10^{-6}\)   & \(10^{-8}\!-\!10^{-2}\)   & Initial grid step (unit of \(y\)).\\
$\gamma$           & 1.167         & \(1.1\!-\!1.2\)           & Geometric expansion factor.\\
$\mathrm{tol}$     & \(10^{-10}\)  & \(10^{-12}\!-\!10^{-6}\)  & Bisection tolerance.\\
\bottomrule
\end{tabular}
\caption{Default constants used by the root-finding module.}
\label{tab:root_params}
\end{table}
%

\begin{algorithm}[h]
\caption{Distribution-aware Conformal Prediction (DCP)}
\label{alg:DCP}
\small
\begin{algorithmic}[1]
\Require Data set $D$; training routine $A$ for the DGP;
         nonconformity score $s(\,\cdot\,)$; nominal miscoverage rate $\alpha$;
         root-finder $r(\,\cdot\,)$
\Ensure Prediction interval $\Bar{\mathcal{C}}_i$ for any new input $x_i^t$

\State \textbf{Split} $D$ into $D^{\text{train}}$, $D^{\text{c}}$ and $D^{\text{t}}$
\State \textbf{Train} DGP: $m \leftarrow A\!\bigl(D^{\text{train}}\bigr)$
\vspace{0.3em}

\Comment{\emph{Calibration phase}}
\ForAll{$(x_i^c,y_i^c)\in D^{\text{c}}$}
    \State $\mathbf{\hat{y}}^{c}_{i} \gets m(x_i^c)$  \Comment{draw vector}
    \State $\varepsilon_i \gets s\bigl(y_i,\mathbf{\hat{y}}^{c}_{i}\bigr)$
\EndFor
\State $\hat q \gets$ empirical $(1-\alpha)$-quantile of $\{\varepsilon_i\}_{i=1}^{N^c}$
\vspace{0.3em}

\Function{Predict}{$x_{i}^t \in D^{t}$}  \Comment{\emph{Inference}}
    \State $\mathbf{\hat{y}}^{t}_{i}
           \gets m(x_{i}^t)$
    \State $f_i(y) \gets s\bigl(y_i,\mathbf{\hat{y}}^{t}_{i}\bigr) - \hat q$
    \State $\mathcal{C}^{\text{low}}_i\!,\, \mathcal{C}^{\text{up}}_i \gets r \bigl(
           f_i(y)\bigr)$
    \State \Return $\mathcal{C}^{\text{low}}_i\!,\, \mathcal{C}^{\text{up}}_i$
\EndFunction
\end{algorithmic}
\end{algorithm}
%

\section{Experimental Setup} 
\label{sec:ExperimentalSetup}
\noindent We evaluate DCP across a diverse collection of univariate time series data sets and three established neural network architectures. Our framework generalizes and unifies several existing state-of-the-art conformal uncertainty methods, such as MC-CP, CMC, CQR and an adaptation of EnbPI, through specific choices of DGP and nonconformity score. By implementing these variants inside a single code base we can benchmark their empirical properties on common data sets and models, isolating the effect of data length, distribution shift, and the balance between aleatoric and epistemic uncertainty. All intervals target a $90\%$ coverage level, using the intended error rate of $\alpha = 0.1$. 

Unless stated otherwise, calibration is performed with the online (sometimes also called adaptive or prequential) variant of split conformal prediction~\citep{Chen2021}. A fixed-length sliding window, initialized with the calibration set, stores the most recent calibration samples. After each newly revealed test target $y^t$ we compute its nonconformity score, append it to the window, discard the oldest element to keep the window length constant, and recompute the global threshold value $\hat{q}$ before the next prediction. This continual update allows the PIs to partially correct for gradual distribution shift. With this, we aim to preserve the finite-sample coverage guarantee and eliminate the need for a separate re-calibration phase. However, for time series with temporal correlations this assumption may still be violated.

\subsection{Data Sets}
\label{sec:datasets}
\noindent We employ two complementary benchmark classes. First, purpose-built \emph{synthetic} series let us isolate the roles of aleatoric versus epistemic uncertainty and therefore highlight the circumstances under which a DGP such as Monte Carlo dropout (MCD) or Quantile Regression (QR) is preferable. Second, six \emph{real-world} collections from the Monash Time Series Repository~\citep{monash2021} expose the framework to a broad spectrum of domains, sampling rates, and dispersion levels, summarized in Table~\ref{tab:dataset_summary}. More details are given in the Supplementary Material A. All experiments share the same pre--processing protocol described at the end of this subsection.

\subsubsection{Synthetic Benchmarks}
\label{sec:synthetic}
\noindent We synthesize two continuous signals of the generic form
\begin{equation}
z_m(t)=\sum_{i=1}^{N} \mathcal{A}_i\sin\bigl(2\pi \omega_i t+\varphi_i\bigr), \quad m=1,2
\end{equation}
with \(\mathcal{A}_i\) denoting the amplitude of the i-th sinusoidal component, \(\omega_i\) its frequency (e.g., day\(^{-1}\)), and \(\varphi_i\) its phase offset (radians). Each signal is sampled at a fixed temporal resolution of 288 samples~day\(^{-1}\) (five-minute spacing) and targets one source of uncertainty.


In the \textit{aleatoric (heteroscedastic-noise) scenario}
the training, validation, and test set share the same clean waveform \((\mathcal{A}_1,\omega_1,\varphi_1)=(1,1,0)\); randomness enters only through heteroscedastic Gaussian noise \(n(t)\sim\mathcal N\!\bigl(0,\sigma^2(t)\bigr)\) with \(\sigma(t)=\sigma_{n}\lvert s(t)\rvert^{\,p}\), $p=0.8$. The noise scale $\sigma_n$ is computed from a target signal-to-noise ratio (SNR) of 15. Due to the amplitude-dependent heteroscedasticity, the effective global SNR is slightly lower. The observed series is then 
\begin{equation} z_{1}(t)=z(t)+n(t). \end{equation}
Because the variance is smallest at the zero crossings and largest near the extrema, QR (which learns conditional quantiles directly) is expected to produce the sharpest \emph{uncalibrated} intervals; after conformalisation, both DGPs regain the same nominal coverage, but we anticipate that the extra distributional information of QR still yields narrower bands than MCD.

In the \textit{epistemic (distribution-shift) scenario}
the training and calibration set again observe only the clean basic waveform \((\mathcal{A}_1,\omega_1,\varphi_1)=(1,1,0)\). The last 10 \% of the series, namely the hold-out test portion, inject two previously unseen harmonics \((\mathcal{A}_2,\omega_2,\varphi_2))=(0.5,0.03,\pi),\;(\mathcal{A}_3,\omega_3,\varphi_1))=(0.15,6,0)\) without measurement noise ($\mathrm{SNR}\to\infty$). Conformal prediction no longer guarantees coverage under this shift, yet a distribution-aware score may still tighten the intervals. Because epistemic uncertainty is better captured by an ensemble of stochastic forward passes, we expect MCD to outperform QR in efficiency.

The two synthetic series last 30 days (aleatoric) and 10 days (epistemic), respectively. Note that online conformalisation is disabled to avoid leaking OOD information into the calibration set. The network architecture, optimizer settings, and draw-vector size are identical across scenarios to isolate the effect of the uncertainty source. Implementation details of the neural architecture and training hyper-parameters are reported in Section \ref{sec:models}.

\begin{table}[ht]
    \centering
    \small 
    
    \begin{threeparttable}
    \begin{tabular}{llccccc}
        \toprule
        \multirow{2}{*}{\textbf{Name}} & 
        \multirow{2}{*}{\textbf{Domain}} & 
        \multirow{2}{*}{\textbf{Freq}} & 
        \multirow{2}{*}{\textbf{N}} & 
        \multirow{2}{*}{\textbf{Var}} & 
        \multicolumn{2}{c}{\textbf{Length}} \\ 
        & & & & & \textbf{Min} & \textbf{Max} \\
        \midrule
        \multirow{2}{*}{M4} & \multirow{2}{*}{multiple} & wk    & 21    & 0.38  & 1019  & 2297 \\
                            &                           & d     & 27    & 0.41  & 3208  & 4454 \\
        Fred MD             & finance                   & mo    & 20    & 1.10  & 728   & 728  \\
        Pedestrian          & mobility                  & d     & 22    & 0.38  & 1009  & 4000 \\    
        Solar               & energy                    & 2h    & 22    & 1.42  & 4380  & 4380 \\
        Wind Farms          & energy                    & h     & 18    & 1.87  & 8784  & 8784 \\
        \bottomrule
    \end{tabular}
    \caption{Summary of Data set Properties}
    \label{tab:dataset_summary}

    \begin{tablenotes}[para, flushleft]
    \footnotesize
    \textit{Note.} Frequency abbreviations: d = daily, wk = weekly, mo = monthly,
    h = hourly. $N$ = number of series. Var = coefficient of variation.
    \end{tablenotes}
    \end{threeparttable}
    
\end{table}

\subsection{Preprocessing}
\noindent Each time series is windowed into X-y (input, target) sample pairs, using a lookback period of three times the forecast horizon which was set to one for all data sets. The resulting samples are then split chronologically into training, validation, and test sets, comprising 70\%, 20\%, and 10\% of the data, respectively. The validation set also serves as the initial calibration set for DCP. A min-max scaler, fitted solely on the corresponding training data, independently normalizes every subsets to the [0, 1] range, thereby preventing data leakage from the validation or test sets.

\subsection{Models}
\label{sec:models}
\noindent We evaluate three neural network architectures commonly used for time series forecasting, namely a Temporal Convolutional Network (TCN)~\citep{colin2016}, a Long Short-Term Memory Network (LSTM)~\citep{hochreiter1997} and a lightweight Temporal Fusion Transformer (TFT)~\citep{LIM20211748}. More details on the model architectures are given in the Supplementary Material B. 

We generate draw vectors in two complementary ways. First, a quantile-regression baseline replaces the standard regression head with a 99-quantile head on an LSTM and trains with the pinball loss. At inference, the set $q_\tau(x)$ for $\tau \in \{0.01,\dots,0.99\}$ is treated as a deterministic pseudo-draw vector (uniform over \(\tau\)), which preserves conformal validity because the same score is used at calibration and inference. This approximation ultimately yields a central $90\%$ prediction interval $[q_{0.05}(x), q_{0.95}(x)]$, while the finite grid truncated tails and monotonicity is not enforced, so crossings may occur.

Second, we obtain stochastic draw vectors via MCD and ensembling. For MCD we keep dropout active at test time and collect 100 forward passes per input from a single trained instance. For Deep Ensembles (DE) and Bootstrap Ensembles (BE), we train 15 independently initialized replicas per architecture. Before each training run we reset the optimizer and any stateful layers, and for DE we reinitialize all network weights with new random seeds. BE trains each replica on a block-bootstrap resample of $D^{\text{train}}$. We draw contiguous 30-step windows with replacement from $\{1,\ldots,N\}$, until the resample contains $\text{r}\times N$ nonunique samples ($r = 0.5$).

\subsection{Nonconformity Scores}
\label{sec:nonconformityscores}
\noindent Although CP is formally distribution--free and delivers finite-sample marginal coverage at level $1-\alpha$ under exchangeability, the choice of a nonconformity score implicitly injects modelling assumptions that determine the width and adaptivity of the PI. In other words, the PI's efficiency hinges on which aspect of the predictor the score exploits. A score that depends only on the error of a point predictor as in \eqref{eq:abs_res}, yields two--sided intervals of constant width for every test point. Such a score does not distinguish between different sources of uncertainty, meaning calibration estimates a single global quantile of the residuals, with no decomposition into aleatoric and epistemic components. When the predictor provides local uncertainty via a predictive distribution (DGP), this information can be embedded in the score. Calibration then estimates a single global quantile $\hat{q}$ of these standardized residuals, making residuals comparable across heteroscedastic inputs. At inference, the interval is \eqref{eq:analy_inverse}, with the unitless $\hat{q}$ being re--scaled by the predictor’s local uncertainty, yielding adaptive widths. In DCP, we exploit this standardize--rescale pairing to combine global calibration with local distributional information.

We investigate three generic score families, each with a different take on uncertainty: (1) Error-based scores, represented by the absolute residual, serve as a symmetric, nonadaptive baseline; (2) Interval-violation scores that measure how far a target falls outside pre-computed bounds such as conditional quantiles or highest-density intervals (HDI), which is useful for heteroscedastic or skewed noise; and (3) Density-based scores, here implemented via a K-Nearest Neighbor density surrogate, which exploit the full shape of possibly multimodal predictive distributions and yield fully data-adaptive prediction sets. Any of these scores can be further normalized by a scalar uncertainty estimate~\citep{angelopoulos2023}, for example, variance for residuals, interval width for quantiles or mean KNN self--distance for density scores. Coverage remains valid as long as the scaling factor depends only on the predictor (and possibly the input $x$) and is strictly positive, ensuring a monotone transformation of the original score. The following subsections formalize the three score families and describe the specific scaling used in our experiments.

\subsubsection{Error-Based Score}
\begin{figure*}[t]
\centering
\includegraphics[width=\textwidth]{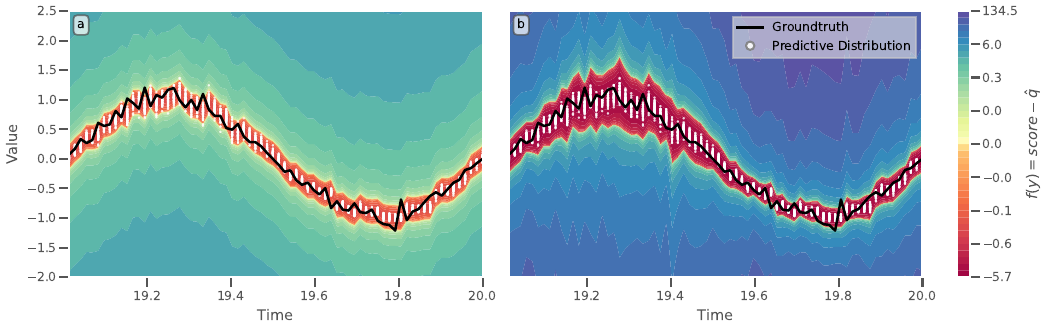}
\caption{Effect of scaled nonconformity scores on the local adaptiveness and thus, sharpness of the prediction interval for (a) the standard residual, and (b) the Z score. For the latter, intervals are narrow or wide where the predictive variance (and thus the predicted uncertainty) is small or high, respectively. Both approaches result in $90\%$ marginal coverage.}
\label{fig:root_func}
\end{figure*}

\noindent The standard residual, the absolute deviation from the predictive mean, is the most straightforward nonconformity score. If no DGP is available, the score reduces to the point forecast error. It is symmetric, completely model-agnostic, and produces nonadaptive intervals of constant width (Fig.~\ref{fig:root_func}a), which makes it a useful baseline. It is defined as
\begin{equation}
\label{eq:residual}
s_{\text{R}}\!\bigl(y,\mathbf{\hat y}^d_i\bigr)
      = \bigl\lvert y - \hat{\mu}_{i}^d\bigr\rvert,
\quad\text{with}\ \, 
\hat{\mu}_{i}^d
      = \frac{1}{M}\sum_{j=1}^{M} \hat y^{d}_{ij}
\end{equation}
\noindent and $M$ is the number of stochastic draws at input $x^d_i$.

An adaptive variant divides the residual in~\eqref{eq:residual} by a local scale derived from the predictive distribution, for example the standard deviation of the draws. The resulting Z-score remains symmetric but widens or narrows with the predictive variance (Fig.~\ref{fig:root_func}b), thereby reflecting the underlying uncertainty regime:
\begin{equation}
\label{eq:res_scaled}
s_{\text{z}}\!\bigl(y,\mathbf{\hat y}^d_i\bigr)
      = \frac{\lvert y - \hat{\mu}_{i}^d\rvert}
              {\sigma_i^d},\\[4pt]
\quad\text{where}\ \, 
\sigma_i^d
      = \sqrt{\frac{1}{M}\sum_{j=1}^{M}
         \bigl(\hat y^{d}_{ij}-\hat{\mu}_{i}^d\bigr)^2}\ .
\end{equation}
\noindent Because the normalization using $\sigma_i^d$ is strictly positive and depends only on the predictor (via $x_i^d$), the conformal guarantee remains valid.

\subsubsection{Interval-Based (Violation) Score}
\label{sec:intervalscore}
\noindent We construct a base inner band from the draw vector $\mathbf{\hat y}^d_i$ with \(\mathcal{C}^{\text{low}}_i\) and \(\mathcal{C}^{\text{up}}_i\) denoting the lower and upper bounds, respectively. We consider two interval-generating statistics: (i) empirical quantiles, in our case the 0.05 and 0.95 quantiles for 90 \% nominal coverage, and (ii) the \((1-\alpha)\) highest-density interval (HDI). The interval-based score is then defined as
\begin{equation}
s_{\text{int}} \bigl(y,\mathbf{\hat y}^d_i\bigr) \;=\;
\max\!\Bigl\{\mathcal{C}^{\text{low}}_i - \;y,\;y\; - \;\mathcal{C}^{\text{up}}_i\Bigr\},
\label{eq:int_base}
\end{equation}
which is negative while $y$ lies inside the base band, zero on the boundary, and positive outside. We intentionally allow $s_{\text{int}}$ to take negative values, so the empirical quantile $\hat{q}$ can be negative; this shrinks the conformal set relative to the base inner band when calibration indicates the band is conservative. The resulting conformal set is~\eqref{eq:int_inverse}, where calibration adds a symmetric slack $\hat{q}$ to the base bounds. Asymmetry in the final PIs arises from the base bounds (e.g., skewed quantiles or HDIs) of the conditional predictive distribution at \(x_i^t\), while the conformal correction itself is symmetric. Because the base bounds $\mathcal{C}^{\text{low}}_i,\mathcal{C}^{\text{up}}_i$ vary with $x_i^t$, the final intervals inherit local uncertainty from the predictive distribution and thus adapt width to the extent the DGP captures the inherent uncertainty. A scaled variant defines the score as the violation divided by the base width, so calibration estimates a global quantile of these normalized violations, and the inversion at inference re-multiplies by the local width to produce adaptive intervals, improving comparability and efficiency across inputs with different noise levels. When the base band is the HDI computed from MCD draws, Eq.~\eqref{eq:int_base} coincides with the violation score used by CMC~\citep{mediyev2025}. The same construction applies to any distributional predictor that provides sample--based bounds, yielding a formally equivalent instantiation of this score family.

\subsubsection{Density-Based Scores}
\label{sec:densityscore}
\noindent A density-based score exploits the full shape of the predictive distribution and does not require the residual to be monotone. Given the draw vector $\mathbf{\hat y}^d_i$, we define the median K-Nearest Neighbor distance using the Euclidean distance metric:
\begin{equation} 
d_{k} \bigl(y,\mathbf{\hat y}^d_i\bigr) \;= \argmin\limits_{d'\in \mathbb{R}_+} \sum_{j=1}^{k} \bigl\lvert d' - |y - \hat{y}^{d}_{i\pi(j)}|\bigr\rvert\, ,
\end{equation}
\noindent with ordering permutation \(\pi\) defined by 
\begin{equation}
|y-\hat{y}^{d}_{i\pi(j)}| \leq |y-\hat{y}^{d}_{i\pi(j+1)}|,\quad \forall j\in \{1,...,M\}     
\end{equation}
\noindent and the normalized nonconformity score as
\begin{equation}
\label{eq:knn_score}
\begin{aligned}
s_{\text{KNN}}\bigl(y,\mathbf{\hat y}^d_i\bigr)
      &= \frac{d_k\!\bigl(y,\mathbf{\hat y}^d_i\bigr)}
              {\tilde d \bigl(\mathbf{\hat y}^d_i\bigr)},\\[4pt]
\text{where}\quad
\tilde d \bigl(\mathbf{\hat y}^d_i\bigr)
      &= \argmin\limits_{d'\in \mathbb{R}_+} \sum_{j,j'=1}^{M}
         \bigl\lvert d' - \lvert\hat y^{d}_{ij'}-\hat y^{d}_{ij}\rvert \bigr\rvert
\end{aligned}
\end{equation}
\noindent is the median of all distances between points in the predictive distribution. We use $k=10$ for all our draw vectors. The denominator depends only on the draws, is strictly positive, and therefore preserves the rank order of the raw distances; the conformal coverage proof is unaffected. 

\subsection{Evaluation Metrics}
\label{sec:evaluation_metrics}
\noindent When producing prediction intervals we evaluate three criteria: (i) \emph{validity}, (ii) \emph{sharpness}, and (iii) \emph{error indication}. The combination of the first two is often referred to as \emph{efficiency}.

Firstly, an interval is considered valid, if its \emph{empirical} coverage \(\widehat C\), measured on a finite test set, does not drop below the \emph{nominal} coverage \((1-\alpha)\), that is, the intervals cover at least \(100(1-\alpha)\%\) of the samples. In the conformal prediction literature this is referred to as \emph{conservative validity}, as opposed to \emph{exact validity}, which requires coverage to equal \(100(1-\alpha)\%\) exactly. Let \(C\) denote the (unknown) \emph{true} coverage the method would achieve; ideally $C \ge 1-\alpha$. In practice only $\widehat C$ is observable, so we compare it with a tolerance-adjusted threshold. For a set of $N^t$ test samples, empirical coverage (also called the prediction interval coverage probability, PICP) is
\begin{equation}
    \widehat C = \frac{1}{N^t} \sum_{i=1}^{N^t} c_{i},
    \quad c_{i} =
    \begin{cases} 
        1, & \text{if } y_{i} \in \left[\mathcal{C}_i^{\text{low}}, \mathcal{C}_i^{\text{up}}\right] \\
        0, & \text{otherwise}
    \end{cases}\ .
    \label{eq:coverage}
\end{equation}
\noindent Treating $\widehat C$ as a binomial sample proportion, its standard error is
\[
\operatorname{SE}(\widehat C)=\sqrt{\frac{C(1-C)}{N^t}}
     \;\approx\;
     \sqrt{\frac{(1-\alpha)\alpha}{N^t}},
\]
where the unknown \(C\) is conservatively replaced by the nominal coverage \(1-\alpha\). Thus, the uncertainty decreases with $1/\sqrt{N^t}$. To accommodate this sampling variability we define a \emph{minimal acceptable coverage} $C_{\text a}$ by subtracting a one-sided \(95\;\%\) confidence margin ($\zeta=1.645$) from the nominal level:
\begin{equation}
C_{\text{a}}(N^t,\alpha,\zeta)
  \;=\;
  1-\alpha
  \;-\;
  \zeta\,\sqrt{\frac{(1-\alpha)\,\alpha}{N^t}}.
\end{equation}
\noindent This avoids unfair rejection of a well-calibrated model evaluated on small test sets, and converges to $1-\alpha$ as \(N^t\) grows.

Secondly, intervals should be sharp. That means the distance between the upper and lower bound of the PI should be as small as possible. A common metric is the Prediction Interval Normalized Average Width (PINAW)~\citep{khosravi2010}
\begin{equation}
    \mathrm{PINAW} = \frac{1}{N^t\cdot \xi} \sum_{i=1}^{N^t}{\delta_i}.
\end{equation}
\noindent Here the interval-wise width \(\delta_i\) is
\begin{equation}
\label{eq:winkler_width}
\delta_i = \mathcal{C}_i^{\text{up}} - \mathcal{C}_i^{\text{low}} 
\end{equation}
and the normalization factor \(\xi\), which makes the metric comparable across different data sets, is defined as the range of the target
\begin{equation}
    \xi \;=\; \max_{1 \le i \le N^t} y_i^t \;-\; \min_{1 \le i \le N^t} y_i^t.
\end{equation}
The adaptivity of intervals can be measured using the coefficient of variance (CV) of the interval width as a simple proxy:
\begin{equation}
\label{eq:cv_delta}
\begin{aligned}
\mu_{\delta}^{} &= \frac{1}{N^t}\sum_{i=1}^{N^t} \delta_i\, ,\\[4pt]
\sigma_\delta^{} &= \sqrt{\frac{1}{N^t-1}\sum_{i=1}^{N^t} \bigl(\delta_i - \mu_{\delta}^{}\bigr)^2}\, , \\[4pt]
\mathrm{CV}_\delta &= \frac{\sigma_\delta^{}}{\mu_{\delta}^{}}.
\end{aligned}
\end{equation}
\noindent CV is unitless and summarizes the relative variability of widths across inputs: $\mathrm{CV}_\delta \approx 0$ indicates near-constant (nonadaptive) intervals, whereas larger values reflect stronger local adaptivity. Note that CV is scale--invariant, so normalizing widths by $\xi$ does not change its value.

\begin{figure}[t]
\centering
\includegraphics[width=\textwidth]{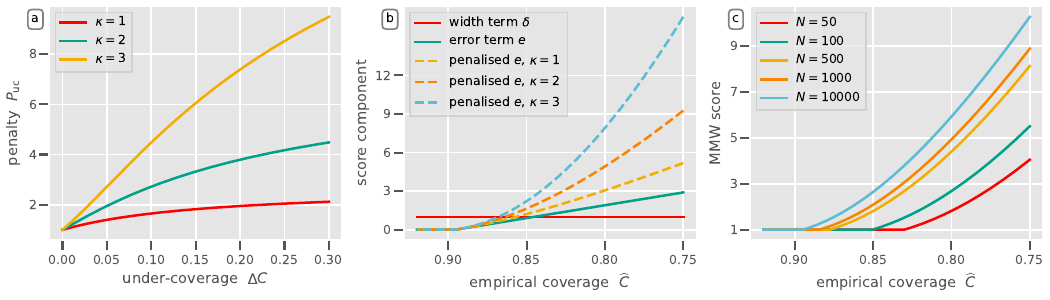}
\caption{Behavior of the MMW score. (a) Under-coverage penalty factor $P_{\text{uc}}$ as a function of the relative under-coverage $\Delta C$ for three different growth factors $\mathrm{K}$. (b) Decomposition of the classical Winkler score into width $\delta$ and error \(e\), together with the penalized error $P_{\text{uc}}\times e$ for three different $\mathrm{K}$ (dashed) of the MMW; all plotted against empirical coverage. (c) MMW score versus empirical coverage for five test-set sizes $N$, where a larger $N$ implies a minimal acceptable coverage $C_a$ closer to $1-\alpha$. All illustrations use $\alpha = 0.1$, $N = 10000$ and $\delta=1$ if not stated otherwise.
}
\label{fig:mmw_score}
\end{figure}

Finally, intervals must be error-indicative, i.e., assign penalties to intervals proportional to the miss distance from the target. In other words, if an interval misses, it should do so by as little as possible so that the distance to the nearest bound remains small and the error is readily apparent. Wider intervals are acceptable only when needed to keep this miss distance minimal. Ideally, a single metric would capture validity, sharpness and error indication. The Coverage Width-Based Criterion (CWC)~\citep{khosravi2010} has been proposed to evaluate PI efficiency. However, it has been strongly criticised on theoretical grounds and is hard to interpret. A closer fit is the \textit{Winkler Interval Score}~\citep{winkler1972, Gneiting01032007}, a \emph{proper} scoring rule that combines interval width and a miss penalty:
\begin{equation}
\label{eq:winkler_main}
W_i = \delta_i + e_i. 
\end{equation}
Here the error term \(\mathrm{e}_i\) is defined as
\begin{equation}
\label{eq:coverage_error}
\mathrm{e}_i = 
\begin{cases} 
\frac{2}{\alpha}\left(\mathcal{C}_i^{\text{low}} - y_i^t \right) & \text{if } y_i^t < \mathcal{C}_i^{\text{low}}, \\
\frac{2}{\alpha} \left(y_i^t - \mathcal{C}_i^{\text{up}} \right) &  \text{if } y_i^t > \mathcal{C}_i^{\text{up}}, \\
0&\text{otherwise}
\end{cases} \, .
\end{equation}
\noindent The width term \(\delta_i\) penalises intervals that are too wide and thus rewards sharpness; its cost grows linearly with slope~\(1\). There is no additional surcharge for mere over-coverage beyond this linear cost.
If an interval \emph{misses}, the error term \(e_i\) adds a distance penalty that also grows linearly, but with slope~\(2/\alpha\). For a typical nominal coverage level $\alpha = 0.1$ the slope becomes $2/\alpha = 20$, so increasing the width by one unit incurs a penalty of 1, whereas missing the target by one unit incurs a penalty of 20. Hence under-coverage is far more expensive per unit distance than over-coverage, yet the growth remains strictly linear: each additional unit of miss adds the same incremental penalty. This is in line with the fact that conformal prediction techniques typically aim for conservatively valid predictors \citep{vovk2022}. In many applications moderate over-coverage is acceptable, whereas under-coverage is critical. We aim to strengthen this behavior. Specifically, we introduce an extra penalty term that amplifies the per-unit miss cost the further the empirical coverage $\widehat C$ falls below the minimal acceptable level $C_{\text a}$ (Fig.~\ref{fig:mmw_score}). The idea is to favor making the interval wider over accepting larger miss distances when under-coverage becomes pronounced. We define an undercoverage penalty as
\begin{equation}
\label{eq:coverage_penalty}
\begin{aligned}
\Delta C &= \max\bigl(0,\; C_{\text{a}} - \widehat C\bigr), \\[4pt]
\rho     &= \frac{\Delta C}{1 - \widehat C}\, , \\[4pt]
P_{\text{uc}} &= e^{\kappa\,\rho}.
\end{aligned}
\end{equation}
\noindent And we obtain the \emph{modified mean Winkler} (MMW) score
\begin{equation}
\text{MMW}
      = \frac{1}{N}\sum_{i=1}^{N}\bigl(\,\delta_i + P_{\text{uc}}\,e_i\bigr)
\label{eq:modified_winkler}
\end{equation}
\noindent for a given set of samples with \(\Delta C\), \(\rho\) and \(\kappa = 2\) denoting under-coverage, relative under-coverage and a growth factor, respectively. The latter value has been choosen to give a reasonably large emphasis on under-coverage.
%
\section{Results and Discussion} 
\label{sec:ResultsAndDiscussion}
\noindent We are not aware of a framework that integrates arbitrary predictor–score pairings with a unified numerical inversion back end for direct comparison, and thus matches DCP in generality. Existing hybrids such as CMC and CQR typically fix a specific predictor--score pair and can be instantiated within DCP as special cases. Because these methods embed different modelling assumptions, direct cross--framework comparisons risk conflating scope with performance. We therefore concentrate on intra--framework comparisons by varying DGPs and nonconformity scores under a shared calibration protocol.

\subsection{DCP under Aleatoric and Epistemic Uncertainty}
\label{sec:res_uncertainty}
\begin{figure*}[t]
\centering
\includegraphics[width=\textwidth]{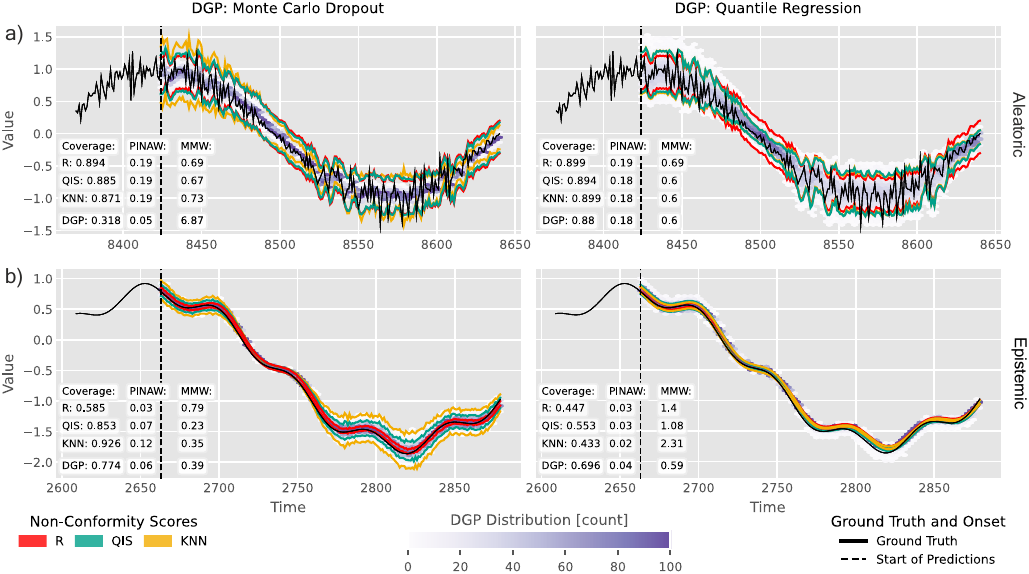}
\caption{Performance of MCD and QR on a) an aleatoric and b) an epistemic synthetic benchmark data set for the LSTM model. The annotated metrics are calculated over the shown test set for 216 samples.}
\label{fig:aleatoric_epistemic}
\end{figure*}

\noindent In the aleatoric (heteroscedastic) scenario (Fig.~\ref{fig:aleatoric_epistemic}a), the model is trained on clean targets while heteroscedastic noise is injected only at test time. MC dropout predominantly reflects epistemic uncertainty or model--sensitivity rather than the true observation noise. This mismatch explains the severe undercoverage of uncalibrated MCD intervals. Conformal calibration restores marginal validity for all scores. The residual baseline (R) does so with uniform--width bands (MMW 0.67--0.73). KNN and QIS remain adaptive under MCD because DCP uses a standardize--rescale scheme. Calibration learns a global quantile on normalized violations. Inference re--scales by a local dispersion proxy $u(x_i^t)$ derived from the predictive draws. With MCD, $u(x_i^t)$ varies with input--dependent model sensitivity (e.g., larger near peaks). As a result, intervals widen where the sine has higher amplitude even though this variation does not reflect aleatoric noise. This yields CVs of 0.39 (KNN) and 0.10 (QIS), with KNN showing more volatile widths due to its noisier density surrogate. In contrast, QR directly learns conditional quantiles and thus encodes the heteroscedastic aleatoric structure. After conformal calibration, only a small slack is needed. The result is the tightest intervals (MMW $\approx$ 0.60 for both KNN and QIS) and consistent adaptivity aligned with the noise pattern (CV = 0.31 for KNN and 0.31 for QIS). Overall, when uncertainty is predominantly aleatoric and heteroscedastic, distributional predictors that estimate conditional spread (e.g., QR) paired with interval--based scores are more efficient and better aligned with the data than MCD--based alternatives. In the latter, adaptivity arises from model sensitivity rather than the true noise structure.

In the epistemic scenario (Fig.~\ref{fig:aleatoric_epistemic}b), we deliberately use an extreme out--of--distribution (OOD) stress test to probe method behavior in limiting scenarios rather than typical operating conditions. The training and calibration data contain only the clean base waveform, while the hold-out test segment introduces previously unseen harmonics without observation noise. This distributional shift breaks exchangeability and uncalibrated intervals are undercovered for both predictors. Point forecasts still track the new regime at a one-step horizon, but the uncertainty signals do not increase enough under shift. In our stres test, conformal calibration restores marginal validity only when the score couples to a local dispersion proxy that responds at test time. With MCD, adaptive scores such as KNN and QIS widen in regions where the predictor’s dispersion is elevated, which improves coverage and maintains efficiency, with MMW values as low as 0.23 for QIS. The residual score underperforms because its constant width is set by a global threshold $\hat{q}$ estimated on in-distribution data, and this underestimates post-shift errors. By contrast, QR does not capture epistemic uncertainty arising from distributional shift. Its base bands remain narrow or misaligned in the OOD segment, none of the QR-based scores achieve valid coverage, and their MMW values are substantially higher than for MCD. This is consistent with the mechanism discussed in the aleatoric case: adaptivity in DCP depends on a local proxy $u(x_i^t)$ that aligns with test--time error, and MCD tends to produce larger $u(x_i^t)$ when the model is more sensitive under shift, whereas QR lacks an epistemic signal. Taken together, the results indicate two requirements for OOD performance. First, a global calibration threshold $\hat{q}$ learned on in-distribution data cannot by itself correct large post-shift errors. Second, adaptivity helps only when the predictive distribution provides a local proxy $u(x_i^t)$ that correlates with error in the shifted regime. In this stress test, MCD-based DCP meets this requirement and QR does not.

Overall, these experiments show that DCP improves efficiency and local adaptivity when the DGP’s uncertainty aligns with the test-time data regime, and that conformal post-processing alone cannot fix a fundamentally uninformative uncertainty signal under shift. In the aleatoric benchmark, QR encodes conditional spread and yields the most efficient calibrated intervals. In the epistemic stress test, MCD produces input-dependent dispersion that supports adaptive widening, whereas QR lacks an epistemic proxy and fails to achieve valid coverage. These results point to practical considerations: choose a predictor whose uncertainty reflects the dominant source in the application, validate conditional coverage and width--uncertainty alignment, and consider sequential or online recalibration when distributional shift is anticipated. Visualizing interval width against the noise proxy or overlaying predictive dispersion and width over time can further clarify how adaptivity manifests in practice.

\subsection{DCP in Practice: Evidence, Guidance, and Limits}
\label{sec:res_dcp}
\noindent Across six benchmarks with diverse sampling rates and sequence lengths, distribution--aware scores generally improve prediction interval efficiency over the nonadaptive residual alternative for all tested models. A detailed summary of empirical coverage and MMW across data sets, models, DGPs, and scores can be found the Supplementary Material C. 

\begin{figure*}[t]
\centering
\includegraphics[width=\textwidth]{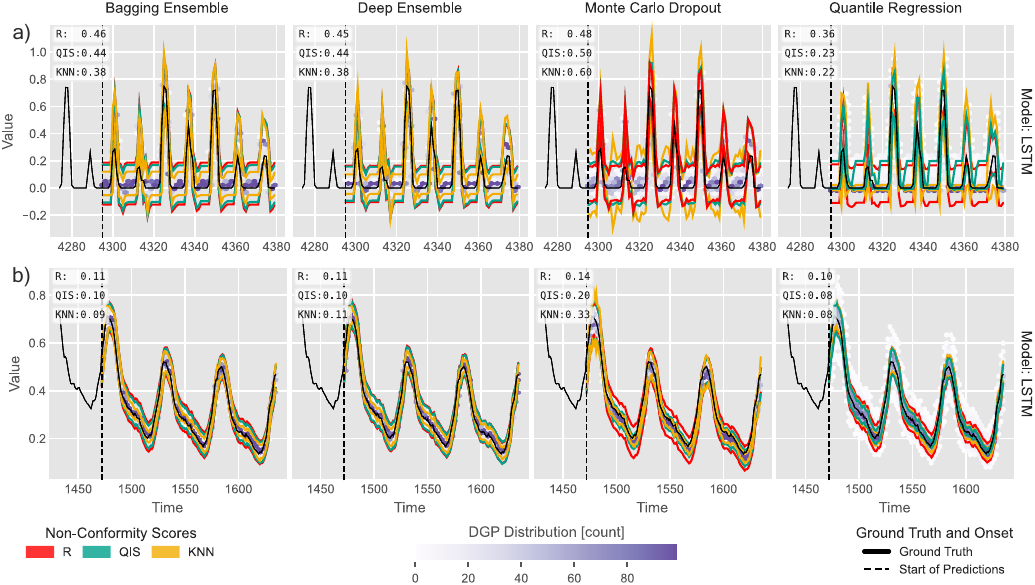}
\caption{DCP intervals on a) Solar (subset 100; limited to 84 test points) and b) M4-Weekly (subset 244) for the LSTM model with all DGPs, combined with residual (R), quantile interval (QIS), and KNN scores. The plot shows normalized ground-truth targets and nominal $90\%$ prediction intervals; annotations report the MMW metric results for each DGP-score pairing.}
\label{fig:solar_weekly}
\end{figure*}

The gains in efficiency are clearest when an informative predictive distribution is coupled with a shape-sensitive score. In other words, the underlying uncertainty is captured by the distribution generating predictor. This is illustrated in the example from the M4--Weekly data set (Fig.~\ref{fig:solar_weekly}b), where adaptive scores (QIS, KNN) improve over the nonadaptive residual baseline when the predictive distribution provides an informative local dispersion signal, reducing interval width and thus MMW at comparable coverage. By contrast, with MCD as the DGP the adaptive scores reflect a misaligned uncertainty signal, yielding over--confident intervals with undercoverage and elevated MMW, underscoring that adaptivity helps only when the dispersion of the predictive distribution aligns with test--time error. 

\begin{figure}[t]
\centering
\includegraphics[width=0.8\textwidth]{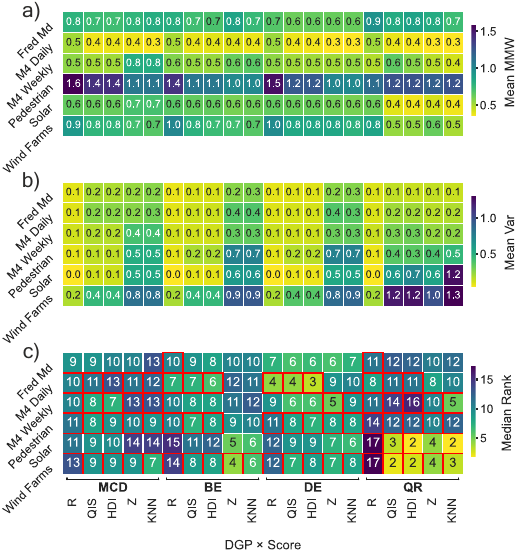}
\caption{Average Performance across all models for the benchmark data sets and all 20 DGP--score combination. Cells values are a) the normalized mean MMW, b) the mean coefficient of variance (CV) of interval width, and c) the median ranks of the MMW score (lower = better), for a given combination, computed over all subsets of each data set. Cells outlined in red mark combinations for which at least \(25 \%\) of the subsets fall below the \(\mathcal{C}_a\) threshold. Note that a CV higher then 0.0 for the nonadaptive residual score is due to online conformal calibration.}
\label{fig:average_performance}
\end{figure}

Furthermore, efficiency hinges on how well data set characteristics are leveraged in the conformalization routine and reflected by the DGP--score combination. For example, on the Solar and Wind Farms data sets, which are nonnegative with many near--zero observations, QR consistently outperforms MCD and ensembles when paired with any adaptive score, reflecting the benefit of scores that exploit the conditional quantile structure in skewed data regimes (Fig.~\ref{fig:average_performance}). Consistent with this, on Solar (Fig.~\ref{fig:solar_weekly}a) QR paired with QIS or KNN produces boundary--aware, adaptive intervals that adhere to nonnegativity and local density, reducing MMW from 0.36 for the nonadaptive baseline to 0.23 and 0.22, respectively. For BE/DE on the other hand, QIS offers limited gains over R while KNN remains modestly adaptive (down to 0.38), illustrating that efficiency gains arise when an informative predictive distribution is coupled with a shape-sensitive score. 

The Pedestrian count data set highlights a failure mode rather than a straightforward efficiency gain. MMW values are among the highest (Fig.~\ref{fig:average_performance}a) and coverage varies across methods, which is consistent with distributional shifts in the COVID--19 period present in the test set. Conformal post--processing does not guarantee coverage under shift, and when the point predictor cannot track the OOD segment, no DGP--score pairing will fully recover validity or efficiency. In our framework, such cases surface as jointly high MMW and volatile or sub--threshold coverage (indicated by the red boxes in Fig.~\ref{fig:average_performance}c), which makes them easy to flag for further analysis and potential shift--aware recalibration. Wind Farms shows a milder version of the same phenomenon. Several methods exhibit near--nominal yet sub--threshold coverage and elevated MMW, consistent with higher volatility and heavier tails in hourly power data.

These outcomes align with our synthetic studies: efficiency hinges on how well the DGP’s uncertainty matches the test--time data regime and on choosing scores that exploit that signal. Under pronounced shift (e.g., Pedestrian), conformal post--processing cannot compensate for an uninformative uncertainty signal, and high MMW coupled with volatile coverage is an indicator to diagnose regime change rather than pursue further tuning.

From a practitioner’s perspective, DCP offers a simple selection rule: first retain only methods that achieve acceptable coverage, then select the combination with the lowest MMW. When the application imposes structural constraints or strong asymmetries (e.g., nonnegativity, skewness), prefer QR as the DGP, best paired with an adaptive score. For comparatively well--specified but noisy series, interval--based scores are a strong default and generally more robust as indicated by the median rank (Fig.~\ref{fig:average_performance}b). Z and KNN score a generally more adaptive than the interval--based scores indicated by their comparatively high coefficient of variance of prediction interval width (Fig.~\ref{fig:average_performance}b). They provide strong adaptivity when the local scale is informative (see Fig.~\ref{fig:root_func}), yet they can show substantial undercoverage or become unstable if the scale estimator is noisy. In these cases, their volatility often leads to less sharp PIs and thus a higher MMW. In settings prone to epistemic shift, ensembles or MCD with adaptive scores are more resilient, but may still require online recalibration and shift diagnostics. Furthermore, we recommend using the MMW as an operational evaluation metric and selector among DGP--score pairings to emphasise undercoverage. It combines sharpness and miss distance in a single number, penalizes undercoverage more steeply only when empirical coverage drops below the minimal acceptable level $C_{a}$, and otherwise behaves like the proper Winkler score. This aligns with decision--maker utility (avoid misses first, then minimise width), discourages jeopardizing coverage for overly narrow intervals, and yields a robust, scalar criterion to tune adaptivity. In other words, when adaptive scores meaningfully improve MMW over the residual baseline, this becomes immediately apparent. PICP and PINAW remain useful companions for reporting coverage and scale--normalized width, but MMW provides the most actionable trade--off for selecting among plug--and--play DGP--score options. The key message is that there is no single combination that works for all regimes; DCP makes it easy to test pairings and adopt what works best for the task at hand.

\subsection{Limitations and Future Directions} 
\label{sec:limits}
\noindent Our bisection-based root-finding algorithm is score-agnostic and numerically lightweight, but its reliability rest on three simple conditions: (i) the score $s(y,\boldsymbol{\hat{y}})$ is finite over the search bracket; (ii) $f(y)$ in Eq.~\ref{eq:root} exhibits at least one sign change within a finite range; and (iii) consecutive crossings are separated enough to be detected on the grid. Heavy tails can push crossings far from the predictive median, and multimodal or step-like score can yield closely spaced or multiple roots. The current retry strategy (adaptive step size and depth) mitigates most cases and additional refinements such as a targeted local fine--grid sweep around detected crossings are left for future work.

Because bounds are obtained by solving Eq.~\ref{eq:root} numerically, constraints imposed by the domain (e.g., nonnegativity, physical limits) can also be alleviated by explicit constrains on the root-finder to improve PI efficiency. For example, restricting the bracket to $[0,\infty)$, allowing one--sided widening, or operating on monotone transforms can be implemented without altering the coverage argument or sacrificing robustness. Another promising extensions to the algorithm is emitting all disjoint interval components, rather than only the outermost, when $f(y)$ has multiple zero crossings. This better reflects genuinely multi--interval predictive sets (e.g., mixture--like regimes) while preserving conformal validity at modest additional cost. Both additions are promising directions that build on DCP’s plug--and--play interface.

Beyond solver--specific considerations, several broader limitations apply. Adaptive scores not only rely on the quality but also the quantity of draws. For example, with small $M$ or an ill-chosen $k/M$ ratio, KNN surrogates can become unstable or overly reactive. In our case, we use $M{=}15$ draws for the ensembles, so $k{=}10$ consumes most of the sampled draws, whereas MCD with $M{=}100$ provides a much smoother density proxy for the same $k$. To stabilise behavior and enable fair comparison across DGPs, choose $k$ relative to $M$ (e.g., $k\!\approx\!0.05$--$0.15\,M$), harmonise $M$ where possible, or bag over several $k$ values. Furthermore, our empirical scope is deliberately constrained. We focus on univariate, one--step--ahead forecasting on a curated set of data sets and subsets. We therefore do not claim an exhaustive taxonomy of uncertainty regimes nor generality beyond these settings. 

We used time series forecasting as a demanding test-bed for DCP because nonstationary data, hard constraints, and mixed uncertainty sources make adaptivity and robustness essential. A natural next step is to extend DGP to multivariate time series and forecasting multiple steps ahead, as it would leverage cross--variable dependencies and propagate uncertainty across horizons, allowing shape--sensitive scores to exploit joint predictive structure. This setting also raises important calibration challenges, such as dependence across variables and horizons and error accumulation, making it a valuable addition to our testbed for distribution--aware conformal methods. Furthermore, mapping DGP--score choices to uncertainty regimes across other inference tasks (e.g., classification or anomaly detection), and to adapt automated selection strategies that pick (or blend) pairings based on quantitative diagnostics would be beneficial in practice. For aleatoric, heteroscedastic regimes, a likelihood--based DGP with a variance head (trained via NLL)~\citep{nix1994, Kendall2017} could be an interesting alternative to QR to supply the local scale that scores exploit within DCP. Finally, a promising avenue for mixed uncertainty regimes is a hybrid DGP that combines aleatoric estimates (e.g., a variance head trained via NLL or QR for conditional quantiles) with epistemic spread (MCD or ensembles). While merging these outputs into a coherent predictive distribution that more accurately quantifies total uncertainty can be nontrivial, the resulting per--input uncertainty can be ingested by DCP so that adaptive nonconformity scores exploit both sources without changes to the pipeline.

\section{Conclusion} 
\noindent We introduced Distribution--aware Conformal Prediction (DCP), a unified framework that couples distribution--generating predictors (DGPs) with score-agnostic conformal calibration to produce valid and efficient prediction intervals for time series. By supporting arbitrary combinations of predictors and nonconformity scores and by obtaining interval bounds through a generic root-finding backend, DCP makes distribution--aware calibration essentially plug--and--play across models, scores, and domains. Together with the modified mean Winkler (MMW) evaluation metric, which jointly accounts for width and undercoverage, the framework offers a practical tool for uncertainty quantification in settings where both validity and efficiency are essential, including high--risk applications subject to emerging regulatory requirements (e.g., the EU AI Act~\citep{EUAIAct2025}).

Our experiments on synthetic and real--world bechmarks yield two main insights. First, interval efficiency is fundamentally constrained by how well the DGP's predictive distribution aligns with the dominant source of uncertainty: when this alignment holds, distribution--aware scores can yield sharper, constraint--aware intervals than nonadaptive residual baselines at comparable coverage. In particular, heteroscedastic (aleatoric) regimes favored quantile--based DGPs paired with interval- or density--aware scores, while regimes dominated by epistemic variability or distribution shift benefit from ensemble or Monte Carlo dropout predictors with adaptive scores. Second, conformal calibration cannot compensate for a poorly trained base predictor or an uninformative uncertainty signal: marginal coverage can be maintained only by inflating bands, in which case point forecasts remain inaccurate and intervals may become either excessively wide or over--confident. In practice, DCP turns method selection into a measurable trade--off: among DGP--score pairings that achieve acceptable coverage, practitioners can simply choose those minimizing the MMW score.

Several limitations and opportunities remain. Temporal dependence violates exchangeability, so our online split--conformal procedure targets approximate marginal coverage and must be validated empirically on each application. Adaptive scores also rely on sufficient draw quality and quantity, and shape--sensitive scores may induce multiple disjoint intervals, which we currently collapse into a single outer interval. Extending DCP to explicitly emit multi--component intervals, to incorporate domain constraints directly in the root--finding routine, and to support hybrid DGPs that combine aleatoric and epistemic components are promising directions. Beyond univariate one–step--ahead forecasting, applying DCP to multivariate and multi--horizon forecasting, as well as to classification, anomaly detection, and spatio--temporal problems, will further test its suitability for modern predictive systems.

Overal, DCP provides a flexible and theoretically grounded starting point for distribution-aware uncertainty quantification in time series analysis. By bridging probabilistic deep learning with rigorous conformal calibration, it enables uncertainty estimates that are not only valid, but also efficient and context-aware, thereby supporting both technical and regulatory requirements in high--stakes decision making.


\acks{The authors used the Fraunhofer GPT (FhGenie) assistant, an internal implementation of the GPT--5 language model, to help with language editing, rephrasing, and brainstorming during manuscript preparation. All technical ideas, problem formulations, algorithms, experiments, and conclusions were developed by the authors.}


\newpage
\appendix




\section{Datasets}
\noindent Our empirical analysis draws on six publicly available time series collections, both real-world and synthetic, from the Monash Time Series Repository which can be found on https://forecastingdata.org/. Two M4 competition subsets provide mixed macro-economic and demographic indicators, while Fred-MD, Solar, Wind Farms and Pedestrian span finance, energy and mobility. All series are treated as uni-variate targets.

The raw data contain no missing values, therefore preprocessing focuses on ensuring consistency across analyzes. We imposed a minimum sequence length for the M4 daily, M4 weekly and pedestrian datasets of 2500, 1000 and 1000 samples, respectively, to exclude excessively short sequences, thereby maintaining consistency and sufficient sample sizes. Furthermore, the solar, wind farms and pedestrian datasets were resampled at two-hourly, hourly and daily intervals, respectively. The chosen frequencies provide a balance between managing data volume and preserving key seasonality and frequency information (e.g., hourly variations in energy consumption and daily cyclic patterns). 

For each dataset we draw 30 random subsets and subsequently discard any subset for which at least two of our models fail to converge under the default hyper--parameter configuration with an acceptable error for the point-forecast on the test set. This keeps the subsequent interval evaluation unbiased and leaves between 18 and 27 valid subsets per dataset. The discard rule prevents excessively wide "trivial" intervals (i.e., marginal coverage is still guaranteed) from poorly fitted models dominating the aggregated evaluation metrics.

\section{Models}
\noindent All models are implemented in TensorFlow 2~\citep{tensorflow2015-whitepaper} and trained with the Adam optimizer~\citep{adam2015} using mean squared error loss, an initial learning rate $10^{-3}$, and gradient clipping with a norm of 1.0. We use a learning-rate schedule that leaves the rate unchanged for the first 35 epochs and then applies an exponential decay by a factor of 0.1 per epoch. Training runs for up to 100 epochs with early stopping after 10 epochs without validation loss improvement, using a batch size of 8. We apply dropout at a rate of 0.2 to hidden dense layers and add recurrent dropout of 0.1 for the LSTM network~\citep{semeniuta2016}. We performed no problem-specific hyper-parameter tuning; the same configuration is used across all datasets and subsets.

The models differ primarily in their sequence-handling blocks, while input projection and output heads are minimally adapted to each architecture. The LSTM uses a single 16-unit-per-direction bidirectional LSTM feeding a 512-unit dense head. The TCN stacks six causal 1D convolutions with kernel size 5 and dilation $2^{0}\ldots 2^{5}$ with residual skip connections, followed by global average pooling to a fixed-length representation. The TFT projects inputs $X \in \mathbb{R}^{B \times L \times 1}$ to a 16-dimensional representation, encodes with a single-layer 16-unit LSTM, and decodes with a single-layer 16-unit LSTM driven by learned horizon-step embeddings $E \in \mathbb{R}^{1 \times 16}$. Decoder states pass through masked causal self-attention with 2 heads and a 16-unit Gated Residual Network. Finally, a time--distributed linear layer yields the 1-step forecast. We adopt this lightweight design because our experiments use a one-step horizon and single-feature inputs, for which the full TFT modules such as variable selection provide limited additional benefit.

\section{Reported Experimental Results for Real-World Benchmark Datasets}
\noindent Tables~\ref{tab:results_summary_a} and~\ref{tab:results_summary_b} summarize empirical coverage and MMW across datasets, models, DGPs, and scores. Note that the MMW values stated in the table are mean and standard deviation of each timeseries collection, with the individual subset MMW being normalized by the test--set range to aid comparison across datasets with different scales and sampling rates. We interpret prediction interval efficiency via the modified mean Winkler (MMW) score, which strongly penalizes under--coverage below the minimal acceptable threshold $C_{\mathrm a}$. Lower MMW therefore reflects both sharper intervals and adequate validity. 

\begin{sidewaystable*}[t]
    \footnotesize
    \renewcommand{\arraystretch}{0.88}
    \setlength{\tabcolsep}{3pt} 
    \centering
    \begin{tabularx}{\textwidth}{L L L Y Y Y Y Y Y Y Y Y Y}
        \toprule
        \multirow{4}{*}{\textbf{Dataset}} 
        & \multirow{4}{*}{\textbf{Model}} 
        & \multirow{4}{*}{\textbf{DGP}} 
        & \multicolumn{4}{c}{\textbf{Error-Based}} 
        & \multicolumn{4}{c}{\textbf{Interval-Based}} 
        & \multicolumn{2}{c}{\textbf{Density-Based}} \\
        \addlinespace
        & &
        & \multicolumn{2}{c}{\textbf{Residuals}} 
        & \multicolumn{2}{c}{\textbf{Z}} 
        & \multicolumn{2}{c}{\textbf{HDI}} 
        & \multicolumn{2}{c}{\textbf{QIS}} 
        & \multicolumn{2}{c}{\textbf{KNN}} \\
        \cmidrule(lr){4-5} \cmidrule(lr){6-7} \cmidrule(lr){8-9} \cmidrule(lr){10-11} \cmidrule(lr){12-13}
        & & 
        & \textbf{C} & \textbf{MMW} 
        & \textbf{C} & \textbf{MMW} 
        & \textbf{C} & \textbf{MMW} 
        & \textbf{C} & \textbf{MMW}
        & \textbf{C} & \textbf{MMW} \\
        \midrule
        \multirow{10}{*}{\shortstack[l]{M4\\Weekly}}
     & \multirow{4}{*}{LSTM}
        & MCD
         & \meanpm{0.90}{0.08} & \meanpm{0.50}{0.27}
         & \meanpm{0.82}{0.12} & \meanpm{0.60}{0.28}
         & \meanpm{0.86}{0.09} & \meanpm{0.48}{0.18}
         & \meanpm{0.86}{0.09} & \meanpm{0.47}{0.17}
         & \meanpm{0.83}{0.11} & \meanpm{0.59}{0.25} \\
     & & BE
         & \meanpm{0.90}{0.08} & \meanpm{0.50}{0.26}
         & \meanpm{0.91}{0.06} & \meanpm{0.46}{0.21}
         & \meanpm{0.91}{0.07} & \meanpm{0.46}{0.21}
         & \meanpm{0.91}{0.07} & \meanpm{0.46}{0.21}
         & \meanpm{0.90}{0.11} & \meanpm{0.52}{0.28} \\
     & & DE
         & \meanpm{0.90}{0.08} & \meanpm{0.48}{0.25}
         & \meanpm{0.90}{0.06} & \meanpm{0.45}{0.24}
         & \meanpm{0.90}{0.07} & \meanpm{0.46}{0.24}
         & \meanpm{0.90}{0.07} & \meanpm{0.46}{0.24}
         & \meanpm{0.89}{0.08} & \meanpm{0.50}{0.30} \\
     & & QR
         & \meanpm{0.90}{0.08} & \meanpm{0.47}{0.24}
         & \meanpm{0.84}{0.07} & \meanpm{0.46}{0.24}
         & \meanpm{0.81}{0.09} & \meanpm{0.55}{0.31}
         & \meanpm{0.79}{0.12} & \meanpm{0.64}{0.44}
         & \meanpm{0.88}{0.06} & \meanpm{\textbf{0.41}}{0.22} \\
      \cmidrule(lr){2-13}
     & \multirow{3}{*}{TCN}
        & MCD
         & \meanpm{0.90}{0.09} & \meanpm{0.56}{0.26}
         & \meanpm{0.75}{0.16} & \meanpm{1.05}{1.17}
         & \meanpm{0.87}{0.07} & \meanpm{0.50}{0.21}
         & \meanpm{0.87}{0.07} & \meanpm{0.50}{0.21}
         & \meanpm{0.75}{0.16} & \meanpm{1.07}{1.16} \\
     & & BE
         & \meanpm{0.90}{0.09} & \meanpm{0.58}{0.25}
         & \meanpm{0.90}{0.11} & \meanpm{0.74}{0.43}
         & \meanpm{0.90}{0.09} & \meanpm{0.56}{0.21}
         & \meanpm{0.89}{0.10} & \meanpm{0.56}{0.21}
         & \meanpm{0.91}{0.11} & \meanpm{0.79}{0.49} \\
     & & DE
         & \meanpm{0.90}{0.09} & \meanpm{0.56}{0.26}
         & \meanpm{0.88}{0.08} & \meanpm{0.57}{0.29}
         & \meanpm{0.89}{0.08} & \meanpm{0.53}{0.23}
         & \meanpm{0.90}{0.08} & \meanpm{\textbf{0.52}}{0.22}
         & \meanpm{0.87}{0.08} & \meanpm{0.57}{0.29} \\
      \cmidrule(lr){2-13}
     & \multirow{3}{*}{TFT}
        & MCD
         & \meanpm{0.90}{0.08} & \meanpm{0.56}{0.23}
         & \meanpm{0.84}{0.12} & \meanpm{0.75}{0.66}
         & \meanpm{0.89}{0.08} & \meanpm{0.53}{0.23}
         & \meanpm{0.89}{0.08} & \meanpm{0.53}{0.22}
         & \meanpm{0.85}{0.11} & \meanpm{0.74}{0.65} \\
     & & BE
         & \meanpm{0.89}{0.11} & \meanpm{0.60}{0.34}
         & \meanpm{0.89}{0.09} & \meanpm{0.64}{0.35}
         & \meanpm{0.90}{0.09} & \meanpm{0.55}{0.24}
         & \meanpm{0.90}{0.09} & \meanpm{0.55}{0.24}
         & \meanpm{0.90}{0.08} & \meanpm{0.61}{0.29} \\
     & & DE
         & \meanpm{0.90}{0.10} & \meanpm{0.57}{0.29}
         & \meanpm{0.91}{0.06} & \meanpm{\textbf{0.48}}{0.20}
         & \meanpm{0.90}{0.09} & \meanpm{0.54}{0.24}
         & \meanpm{0.90}{0.09} & \meanpm{0.54}{0.24}
         & \meanpm{0.91}{0.07} & \meanpm{0.50}{0.21} \\
         \midrule
                  
    \multirow{10}{*}{\shortstack[l]{M4\\Daily}}
     & \multirow{4}{*}{LSTM}
        & MCD
         & \meanpm{0.85}{0.13} & \meanpm{0.51}{0.91}
         & \meanpm{0.89}{0.10} & \meanpm{0.37}{0.46}
         & \meanpm{0.89}{0.09} & \meanpm{0.34}{0.33}
         & \meanpm{0.89}{0.10} & \meanpm{0.34}{0.35}
         & \meanpm{0.90}{0.09} & \meanpm{0.34}{0.31} \\
     & & BE
         & \meanpm{0.88}{0.08} & \meanpm{0.34}{0.52}
         & \meanpm{0.93}{0.06} & \meanpm{0.33}{0.23}
         & \meanpm{0.90}{0.06} & \meanpm{0.27}{0.28}
         & \meanpm{0.90}{0.06} & \meanpm{0.28}{0.31}
         & \meanpm{0.93}{0.06} & \meanpm{0.33}{0.21} \\
     & & DE
         & \meanpm{0.87}{0.10} & \meanpm{0.35}{0.55}
         & \meanpm{0.93}{0.06} & \meanpm{\textbf{0.26}}{0.15}
         & \meanpm{0.89}{0.08} & \meanpm{0.27}{0.27}
         & \meanpm{0.89}{0.08} & \meanpm{0.27}{0.25}
         & \meanpm{0.93}{0.06} & \meanpm{0.29}{0.17} \\
     & & QR
         & \meanpm{0.84}{0.15} & \meanpm{0.50}{0.76}
         & \meanpm{0.87}{0.12} & \meanpm{0.33}{0.35}
         & \meanpm{0.88}{0.12} & \meanpm{0.35}{0.33}
         & \meanpm{0.88}{0.13} & \meanpm{0.37}{0.37}
         & \meanpm{0.89}{0.11} & \meanpm{0.33}{0.32} \\
      \cmidrule(lr){2-13}
     & \multirow{3}{*}{TCN}
        & MCD
         & \meanpm{0.89}{0.09} & \meanpm{0.37}{0.70}
         & \meanpm{0.91}{0.07} & \meanpm{0.32}{0.41}
         & \meanpm{0.91}{0.07} & \meanpm{0.30}{0.18}
         & \meanpm{0.91}{0.07} & \meanpm{0.29}{0.16}
         & \meanpm{0.92}{0.06} & \meanpm{0.31}{0.25} \\
     & & BE
         & \meanpm{0.89}{0.08} & \meanpm{0.35}{0.58}
         & \meanpm{0.93}{0.06} & \meanpm{0.34}{0.26}
         & \meanpm{0.91}{0.05} & \meanpm{0.27}{0.28}
         & \meanpm{0.91}{0.05} & \meanpm{0.26}{0.22}
         & \meanpm{0.94}{0.06} & \meanpm{0.33}{0.24} \\
     & & DE
         & \meanpm{0.89}{0.09} & \meanpm{0.34}{0.56}
         & \meanpm{0.94}{0.06} & \meanpm{0.35}{0.38}
         & \meanpm{0.91}{0.06} & \meanpm{\textbf{0.24}}{0.14}
         & \meanpm{0.91}{0.06} & \meanpm{\textbf{0.24}}{0.14}
         & \meanpm{0.94}{0.07} & \meanpm{0.35}{0.39} \\
      \cmidrule(lr){2-13}
     & \multirow{3}{*}{TFT}
        & MCD
         & \meanpm{0.81}{0.17} & \meanpm{0.62}{0.72}
         & \meanpm{0.88}{0.09} & \meanpm{0.39}{0.40}
         & \meanpm{0.84}{0.12} & \meanpm{0.47}{0.49}
         & \meanpm{0.84}{0.12} & \meanpm{0.45}{0.46}
         & \meanpm{0.89}{0.08} & \meanpm{0.38}{0.33} \\
     & & BE
         & \meanpm{0.81}{0.19} & \meanpm{0.78}{1.28}
         & \meanpm{0.87}{0.17} & \meanpm{0.51}{0.82}
         & \meanpm{0.84}{0.18} & \meanpm{0.61}{1.20}
         & \meanpm{0.84}{0.18} & \meanpm{0.59}{1.09}
         & \meanpm{0.90}{0.14} & \meanpm{0.45}{0.76} \\
     & & DE
         & \meanpm{0.80}{0.22} & \meanpm{0.86}{1.46}
         & \meanpm{0.89}{0.13} & \meanpm{0.38}{0.29}
         & \meanpm{0.83}{0.17} & \meanpm{0.55}{0.77}
         & \meanpm{0.83}{0.17} & \meanpm{0.58}{0.85}
         & \meanpm{0.91}{0.10} & \meanpm{\textbf{0.35}}{0.27} \\
         \midrule

    \multirow{10}{*}{\shortstack[l]{Fred\\MD}}
     & \multirow{4}{*}{LSTM}
        & MCD
         & \meanpm{0.87}{0.20} & \meanpm{0.80}{1.02}
         & \meanpm{0.90}{0.14} & \meanpm{0.78}{0.61}
         & \meanpm{0.91}{0.13} & \meanpm{0.74}{0.67}
         & \meanpm{0.90}{0.15} & \meanpm{0.74}{0.71}
         & \meanpm{0.93}{0.12} & \meanpm{0.82}{0.58} \\
     & & BE
         & \meanpm{0.86}{0.24} & \meanpm{0.75}{0.70}
         & \meanpm{0.89}{0.20} & \meanpm{0.75}{0.51}
         & \meanpm{0.89}{0.20} & \meanpm{0.59}{0.41}
         & \meanpm{0.89}{0.21} & \meanpm{0.61}{0.45}
         & \meanpm{0.91}{0.13} & \meanpm{0.66}{0.37} \\
     & & DE
         & \meanpm{0.85}{0.24} & \meanpm{0.75}{0.77}
         & \meanpm{0.88}{0.16} & \meanpm{0.61}{0.42}
         & \meanpm{0.86}{0.22} & \meanpm{0.70}{0.67}
         & \meanpm{0.85}{0.23} & \meanpm{0.70}{0.67}
         & \meanpm{0.88}{0.13} & \meanpm{\textbf{0.58}}{0.32} \\
     & & QR
         & \meanpm{0.83}{0.27} & \meanpm{0.94}{1.28}
         & \meanpm{0.85}{0.24} & \meanpm{0.81}{0.84}
         & \meanpm{0.85}{0.25} & \meanpm{0.85}{0.97}
         & \meanpm{0.85}{0.24} & \meanpm{0.81}{0.81}
         & \meanpm{0.91}{0.14} & \meanpm{0.72}{0.64} \\
      \cmidrule(lr){2-13}
     & \multirow{3}{*}{TCN}
        & MCD
         & \meanpm{0.92}{0.07} & \meanpm{0.57}{0.26}
         & \meanpm{0.94}{0.06} & \meanpm{0.70}{0.49}
         & \meanpm{0.94}{0.06} & \meanpm{0.66}{0.29}
         & \meanpm{0.95}{0.06} & \meanpm{0.63}{0.28}
         & \meanpm{0.94}{0.06} & \meanpm{0.77}{0.46} \\
     & & BE
         & \meanpm{0.92}{0.07} & \meanpm{0.56}{0.34}
         & \meanpm{0.92}{0.12} & \meanpm{0.77}{0.53}
         & \meanpm{0.93}{0.07} & \meanpm{0.55}{0.31}
         & \meanpm{0.93}{0.07} & \meanpm{0.55}{0.32}
         & \meanpm{0.93}{0.09} & \meanpm{0.71}{0.47} \\
     & & DE
         & \meanpm{0.92}{0.08} & \meanpm{0.56}{0.31}
         & \meanpm{0.93}{0.07} & \meanpm{0.55}{0.32}
         & \meanpm{0.93}{0.07} & \meanpm{\textbf{0.53}}{0.29}
         & \meanpm{0.93}{0.07} & \meanpm{\textbf{0.53}}{0.29}
         & \meanpm{0.94}{0.07} & \meanpm{\textbf{0.53}}{0.29} \\
      \cmidrule(lr){2-13}
     & \multirow{3}{*}{TFT}
        & MCD
         & \meanpm{0.83}{0.24} & \meanpm{0.92}{0.72}
         & \meanpm{0.87}{0.22} & \meanpm{0.78}{0.50}
         & \meanpm{0.89}{0.14} & \meanpm{0.73}{0.30}
         & \meanpm{0.87}{0.18} & \meanpm{0.81}{0.40}
         & \meanpm{0.95}{0.06} & \meanpm{0.79}{0.49} \\
     & & BE
         & \meanpm{0.84}{0.19} & \meanpm{1.05}{0.66}
         & \meanpm{0.88}{0.14} & \meanpm{1.01}{0.59}
         & \meanpm{0.90}{0.12} & \meanpm{0.90}{0.50}
         & \meanpm{0.89}{0.10} & \meanpm{0.96}{0.53}
         & \meanpm{0.91}{0.10} & \meanpm{0.82}{0.47} \\
     & & DE
         & \meanpm{0.90}{0.12} & \meanpm{0.67}{0.45}
         & \meanpm{0.93}{0.08} & \meanpm{0.69}{0.37}
         & \meanpm{0.93}{0.08} & \meanpm{0.63}{0.35}
         & \meanpm{0.94}{0.06} & \meanpm{\textbf{0.58}}{0.30}
         & \meanpm{0.91}{0.13} & \meanpm{0.75}{0.40} \\

\bottomrule
\end{tabularx}
\caption{DCP Results Across Datasets, Models, DGPs and Scores}
\label{tab:results_summary_a}
\end{sidewaystable*}

\vfill

\begin{sidewaystable*}[t]
    \footnotesize
    \renewcommand{\arraystretch}{0.88}
    \setlength{\tabcolsep}{3pt} 
    \centering
    \begin{tabularx}{\textwidth}{L L L Y Y Y Y Y Y Y Y Y Y}
        \toprule
        \multirow{4}{*}{\textbf{Dataset}} 
        & \multirow{4}{*}{\textbf{Model}} 
        & \multirow{4}{*}{\textbf{DGP}} 
        & \multicolumn{4}{c}{\textbf{Error-Based}} 
        & \multicolumn{4}{c}{\textbf{Interval-Based}} 
        & \multicolumn{2}{c}{\textbf{Density-Based}} \\
        \addlinespace
        & &
        & \multicolumn{2}{c}{\textbf{Residuals}} 
        & \multicolumn{2}{c}{\textbf{Z}} 
        & \multicolumn{2}{c}{\textbf{HDI}} 
        & \multicolumn{2}{c}{\textbf{QIS}} 
        & \multicolumn{2}{c}{\textbf{KNN}} \\
        \cmidrule(lr){4-5} \cmidrule(lr){6-7} \cmidrule(lr){8-9} \cmidrule(lr){10-11} \cmidrule(lr){12-13}
        & & 
        & \textbf{C} & \textbf{MMW} 
        & \textbf{C} & \textbf{MMW} 
        & \textbf{C} & \textbf{MMW} 
        & \textbf{C} & \textbf{MMW}
        & \textbf{C} & \textbf{MMW} \\
        \midrule
    \multirow{10}{*}{\shortstack[l]{Pedestrian\\Count}}
     & \multirow{4}{*}{LSTM}
        & MCD
         & \meanpm{0.84}{0.08} & \meanpm{0.98}{0.90}
         & \meanpm{0.88}{0.07} & \meanpm{0.89}{0.47}
         & \meanpm{0.86}{0.07} & \meanpm{0.88}{0.76}
         & \meanpm{0.86}{0.07} & \meanpm{0.87}{0.74}
         & \meanpm{0.88}{0.07} & \meanpm{0.89}{0.45} \\
     & & BE
         & \meanpm{0.84}{0.08} & \meanpm{0.99}{0.77}
         & \meanpm{0.91}{0.04} & \meanpm{0.95}{0.41}
         & \meanpm{0.87}{0.08} & \meanpm{0.83}{0.58}
         & \meanpm{0.86}{0.08} & \meanpm{0.83}{0.56}
         & \meanpm{0.91}{0.03} & \meanpm{0.97}{0.45} \\
     & & DE
         & \meanpm{0.84}{0.09} & \meanpm{0.98}{0.74}
         & \meanpm{0.90}{0.05} & \meanpm{\textbf{0.80}}{0.27}
         & \meanpm{0.85}{0.08} & \meanpm{0.90}{0.66}
         & \meanpm{0.85}{0.08} & \meanpm{0.90}{0.66}
         & \meanpm{0.90}{0.05} & \meanpm{0.81}{0.27} \\
     & & QR
         & \meanpm{0.83}{0.09} & \meanpm{1.13}{0.90}
         & \meanpm{0.82}{0.08} & \meanpm{1.19}{1.19}
         & \meanpm{0.82}{0.09} & \meanpm{1.25}{1.30}
         & \meanpm{0.81}{0.08} & \meanpm{1.24}{1.25}
         & \meanpm{0.82}{0.09} & \meanpm{1.24}{1.25} \\
      \cmidrule(lr){2-13}
     & \multirow{3}{*}{TCN}
        & MCD
         & \meanpm{0.81}{0.10} & \meanpm{2.26}{2.90}
         & \meanpm{0.85}{0.09} & \meanpm{1.13}{0.67}
         & \meanpm{0.82}{0.09} & \meanpm{1.89}{2.23}
         & \meanpm{0.82}{0.09} & \meanpm{1.88}{2.20}
         & \meanpm{0.84}{0.09} & \meanpm{1.17}{0.73} \\
     & & BE
         & \meanpm{0.81}{0.10} & \meanpm{1.67}{1.42}
         & \meanpm{0.90}{0.05} & \meanpm{1.12}{0.53}
         & \meanpm{0.83}{0.09} & \meanpm{\textbf{1.08}}{0.72}
         & \meanpm{0.83}{0.09} & \meanpm{\textbf{1.08}}{0.69}
         & \meanpm{0.89}{0.06} & \meanpm{1.12}{0.51} \\
     & & DE
         & \meanpm{0.81}{0.10} & \meanpm{2.13}{2.25}
         & \meanpm{0.88}{0.07} & \meanpm{1.23}{0.68}
         & \meanpm{0.83}{0.10} & \meanpm{1.40}{1.33}
         & \meanpm{0.83}{0.09} & \meanpm{1.42}{1.30}
         & \meanpm{0.88}{0.06} & \meanpm{1.17}{0.63} \\
      \cmidrule(lr){2-13}
     & \multirow{3}{*}{TFT}
        & MCD
         & \meanpm{0.80}{0.10} & \meanpm{1.44}{0.86}
         & \meanpm{0.85}{0.10} & \meanpm{1.17}{0.80}
         & \meanpm{0.80}{0.10} & \meanpm{1.40}{0.85}
         & \meanpm{0.80}{0.10} & \meanpm{1.40}{0.85}
         & \meanpm{0.85}{0.10} & \meanpm{1.17}{0.79} \\
     & & BE
         & \meanpm{0.80}{0.09} & \meanpm{1.46}{0.89}
         & \meanpm{0.90}{0.06} & \meanpm{\textbf{0.86}}{0.33}
         & \meanpm{0.80}{0.09} & \meanpm{1.30}{0.71}
         & \meanpm{0.80}{0.09} & \meanpm{1.31}{0.74}
         & \meanpm{0.90}{0.05} & \meanpm{0.87}{0.36} \\
     & & DE
         & \meanpm{0.80}{0.09} & \meanpm{1.41}{0.81}
         & \meanpm{0.89}{0.07} & \meanpm{0.96}{0.35}
         & \meanpm{0.80}{0.09} & \meanpm{1.33}{0.74}
         & \meanpm{0.80}{0.09} & \meanpm{1.33}{0.73}
         & \meanpm{0.89}{0.07} & \meanpm{0.96}{0.39} \\
         \midrule
         
        \multirow{10}{*}{\shortstack[l]{Solar}}
     & \multirow{4}{*}{LSTM}
        & MCD
         & \meanpm{0.90}{0.01} & \meanpm{0.58}{0.08}
         & \meanpm{0.89}{0.01} & \meanpm{0.66}{0.07}
         & \meanpm{0.90}{0.01} & \meanpm{0.59}{0.08}
         & \meanpm{0.90}{0.01} & \meanpm{0.59}{0.08}
         & \meanpm{0.89}{0.01} & \meanpm{0.67}{0.08} \\
     & & BE
         & \meanpm{0.90}{0.01} & \meanpm{0.62}{0.08}
         & \meanpm{0.92}{0.02} & \meanpm{0.47}{0.09}
         & \meanpm{0.90}{0.01} & \meanpm{0.58}{0.08}
         & \meanpm{0.90}{0.01} & \meanpm{0.58}{0.08}
         & \meanpm{0.92}{0.03} & \meanpm{0.49}{0.10} \\
     & & DE
         & \meanpm{0.90}{0.01} & \meanpm{0.58}{0.08}
         & \meanpm{0.91}{0.03} & \meanpm{0.51}{0.09}
         & \meanpm{0.90}{0.01} & \meanpm{0.56}{0.08}
         & \meanpm{0.90}{0.01} & \meanpm{0.56}{0.08}
         & \meanpm{0.90}{0.03} & \meanpm{0.53}{0.09} \\
     & & QR
         & \meanpm{0.90}{0.01} & \meanpm{0.60}{0.08}
         & \meanpm{0.90}{0.02} & \meanpm{0.41}{0.07}
         & \meanpm{0.90}{0.02} & \meanpm{\textbf{0.36}}{0.07}
         & \meanpm{0.88}{0.05} & \meanpm{0.39}{0.07}
         & \meanpm{0.91}{0.01} & \meanpm{\textbf{0.36}}{0.05} \\
      \cmidrule(lr){2-13}
     & \multirow{3}{*}{TCN}
        & MCD
         & \meanpm{0.91}{0.02} & \meanpm{0.55}{0.07}
         & \meanpm{0.91}{0.01} & \meanpm{0.46}{0.06}
         & \meanpm{0.91}{0.01} & \meanpm{0.51}{0.07}
         & \meanpm{0.91}{0.02} & \meanpm{0.51}{0.07}
         & \meanpm{0.91}{0.01} & \meanpm{0.46}{0.06} \\
     & & BE
         & \meanpm{0.90}{0.02} & \meanpm{0.59}{0.08}
         & \meanpm{0.93}{0.02} & \meanpm{\textbf{0.45}}{0.07}
         & \meanpm{0.91}{0.02} & \meanpm{0.55}{0.08}
         & \meanpm{0.91}{0.02} & \meanpm{0.55}{0.08}
         & \meanpm{0.93}{0.02} & \meanpm{\textbf{0.45}}{0.07} \\
     & & DE
         & \meanpm{0.91}{0.01} & \meanpm{0.56}{0.08}
         & \meanpm{0.93}{0.02} & \meanpm{0.46}{0.08}
         & \meanpm{0.92}{0.01} & \meanpm{0.53}{0.09}
         & \meanpm{0.92}{0.01} & \meanpm{0.53}{0.08}
         & \meanpm{0.93}{0.02} & \meanpm{0.46}{0.08} \\
      \cmidrule(lr){2-13}
     & \multirow{3}{*}{TFT}
        & MCD
         & \meanpm{0.90}{0.01} & \meanpm{0.75}{0.07}
         & \meanpm{0.89}{0.01} & \meanpm{0.95}{0.11}
         & \meanpm{0.90}{0.01} & \meanpm{0.75}{0.07}
         & \meanpm{0.90}{0.01} & \meanpm{0.75}{0.07}
         & \meanpm{0.89}{0.01} & \meanpm{0.95}{0.11} \\
     & & BE
         & \meanpm{0.90}{0.01} & \meanpm{\textbf{0.74}}{0.08}
         & \meanpm{0.90}{0.02} & \meanpm{0.83}{0.10}
         & \meanpm{0.90}{0.01} & \meanpm{\textbf{0.74}}{0.08}
         & \meanpm{0.90}{0.01} & \meanpm{\textbf{0.74}}{0.08}
         & \meanpm{0.89}{0.05} & \meanpm{0.86}{0.21} \\
     & & DE
         & \meanpm{0.90}{0.01} & \meanpm{\textbf{0.74}}{0.08}
         & \meanpm{0.90}{0.04} & \meanpm{0.80}{0.16}
         & \meanpm{0.90}{0.01} & \meanpm{\textbf{0.74}}{0.08}
         & \meanpm{0.90}{0.01} & \meanpm{\textbf{0.74}}{0.08}
         & \meanpm{0.90}{0.03} & \meanpm{0.76}{0.11} \\
         \midrule
         
    \multirow{10}{*}{\shortstack[l]{Wind\\Farms}}
     & \multirow{4}{*}{LSTM}
        & MCD
         & \meanpm{0.86}{0.11} & \meanpm{0.84}{1.08}
         & \meanpm{0.87}{0.09} & \meanpm{0.63}{0.53}
         & \meanpm{0.87}{0.09} & \meanpm{0.69}{0.75}
         & \meanpm{0.87}{0.09} & \meanpm{0.69}{0.73}
         & \meanpm{0.88}{0.08} & \meanpm{0.63}{0.50} \\
     & & BE
         & \meanpm{0.83}{0.21} & \meanpm{0.95}{1.31}
         & \meanpm{0.84}{0.23} & \meanpm{0.74}{1.00}
         & \meanpm{0.87}{0.10} & \meanpm{0.71}{0.76}
         & \meanpm{0.84}{0.14} & \meanpm{0.78}{0.79}
         & \meanpm{0.88}{0.12} & \meanpm{0.64}{0.65} \\
     & & DE
         & \meanpm{0.83}{0.21} & \meanpm{0.93}{1.30}
         & \meanpm{0.84}{0.20} & \meanpm{0.72}{0.91}
         & \meanpm{0.86}{0.13} & \meanpm{0.79}{0.92}
         & \meanpm{0.86}{0.13} & \meanpm{0.79}{0.92}
         & \meanpm{0.86}{0.11} & \meanpm{0.68}{0.62} \\
     & & QR
         & \meanpm{0.85}{0.15} & \meanpm{0.84}{1.10}
         & \meanpm{0.86}{0.08} & \meanpm{0.62}{0.64}
         & \meanpm{0.82}{0.21} & \meanpm{0.53}{0.54}
         & \meanpm{0.82}{0.21} & \meanpm{0.53}{0.54}
         & \meanpm{0.88}{0.05} & \meanpm{\textbf{0.49}}{0.30} \\
      \cmidrule(lr){2-13}
     & \multirow{3}{*}{TCN}
        & MCD
         & \meanpm{0.87}{0.11} & \meanpm{0.89}{1.09}
         & \meanpm{0.87}{0.08} & \meanpm{0.75}{0.72}
         & \meanpm{0.88}{0.08} & \meanpm{0.74}{0.79}
         & \meanpm{0.88}{0.08} & \meanpm{0.75}{0.80}
         & \meanpm{0.88}{0.07} & \meanpm{0.70}{0.63} \\
     & & BE
         & \meanpm{0.85}{0.14} & \meanpm{0.98}{1.27}
         & \meanpm{0.87}{0.14} & \meanpm{0.67}{0.72}
         & \meanpm{0.88}{0.08} & \meanpm{0.69}{0.83}
         & \meanpm{0.88}{0.08} & \meanpm{0.69}{0.82}
         & \meanpm{0.87}{0.12} & \meanpm{\textbf{0.66}}{0.67} \\
     & & DE
         & \meanpm{0.86}{0.11} & \meanpm{0.92}{1.17}
         & \meanpm{0.85}{0.18} & \meanpm{0.79}{0.97}
         & \meanpm{0.88}{0.09} & \meanpm{0.72}{0.87}
         & \meanpm{0.87}{0.12} & \meanpm{0.74}{0.90}
         & \meanpm{0.86}{0.14} & \meanpm{0.74}{0.82} \\
      \cmidrule(lr){2-13}
     & \multirow{3}{*}{TFT}
        & MCD
         & \meanpm{0.86}{0.11} & \meanpm{0.97}{1.24}
         & \meanpm{0.88}{0.08} & \meanpm{0.71}{0.77}
         & \meanpm{0.87}{0.10} & \meanpm{0.84}{0.97}
         & \meanpm{0.87}{0.10} & \meanpm{0.84}{0.97}
         & \meanpm{0.89}{0.05} & \meanpm{\textbf{0.64}}{0.53} \\
     & & BE
         & \meanpm{0.83}{0.15} & \meanpm{1.04}{1.32}
         & \meanpm{0.86}{0.13} & \meanpm{0.70}{0.60}
         & \meanpm{0.86}{0.10} & \meanpm{0.77}{0.84}
         & \meanpm{0.84}{0.17} & \meanpm{0.84}{0.95}
         & \meanpm{0.87}{0.10} & \meanpm{0.66}{0.52} \\
     & & DE
         & \meanpm{0.82}{0.22} & \meanpm{1.05}{1.53}
         & \meanpm{0.79}{0.22} & \meanpm{0.99}{1.27}
         & \meanpm{0.81}{0.22} & \meanpm{0.97}{1.32}
         & \meanpm{0.82}{0.22} & \meanpm{0.93}{1.32}
         & \meanpm{0.83}{0.15} & \meanpm{0.88}{0.88} \\
\bottomrule
\end{tabularx}
\caption{DCP Results Across Datasets, Models, DGPs and Scores}
\label{tab:results_summary_b}
\end{sidewaystable*}

\vfill
\clearpage
\newpage

\section{Nomenclature}

\begingroup
\footnotesize 
\setlist[description]{style=multiline,labelwidth=2.4cm,itemsep=3pt,leftmargin=2.6cm}
\begin{description}
  \item[$D$] Dataset
  \item[$D^{\mathrm{train}}, D^c, D^t$] training, calibration and test set
  \item[$N^c, N^t$] number of samples in $D^c$ and $D^t$, respectively
  \item[$(x,y) \in D$] data point with input $x$ and ground truth $y$
  \item[$x_i^d$] $i$-th input sample from $D^d$
  \item[$y_i^d$] ground truth value of $i$-th data sample from $D^d$
  \item[$\hat{P}_x$] predictive distribution
  \item[$\mathbf{\hat{y}}(x) \subseteq \hat{P}_x$] draw vector
  \item[$\mathbf{\hat{y}}_i^d$] draw vector of sample $i$ from $D^d$
  \item[$M$] number of draws
  \item[$\hat \mu (x)$] point prediction or mean value of $\boldsymbol{\hat{y}}(x)$
  \item[$\tilde{y}(x)$] predictive median
  \item[$\theta$] quantile level
  \item[$\alpha$] nominal miscoverage rate
  \item[$\hat{q}$] $100(1-\alpha)\%$-quantile of conformal scores
  \item[$s$] nonconformity score function
  \item[$\Bar{\mathcal{C}}$] split conformal prediction interval
  \item[$\mathcal{C}$] predictive set
  \item[$\mathcal{C}^{\mathrm{low}}_i$, $\mathcal{C}^{\mathrm{up}}_i$] interval boundaries
  \item[$m$] trained DGP
  \item[$\mathrm{A}$] training routine
  \item[$\varepsilon$] nonconformity score value
  \item[$u$] uncertainty scalar
  \item[$r$] root finder
  \item[$\mathrm{depth}$] Number of draws per query
  \item[$h_0$] Initial grid step (unit of $y$)
  \item[$\gamma$] Geometric expansion factor
  \item[$\mathrm{tol}$] Tolerance of the bisection routine
  \item[$z$] continuous signal for synthetic data sets
  \item[$\mathcal{A}$] amplitude of sinusoidal component
  \item[$\omega$] frequency (e.g., day$^{-1}$)
  \item[$\varphi$] phase offset (radians)
  \item[$s_{\mathrm{R}}$] Residual score function
  \item[$s_{\mathrm{Z}}$] Z score function
  \item[$s_{\mathrm{int}}$] Interval--based score function
  \item[$s_{\mathrm{KNN}}$] KNN score function
  \item[$C$] true coverage
  \item[$\widehat C$] empirical coverage
  \item[$C_a$] minimal acceptable coverage
  \item[$W$] Winkler Interval Score
  \item[$\delta$] interval width
  \item[$\mathrm{CV}_\delta$] coefficient of variance of the interval width
  \item[$\mathrm{MMW}$] Modified Mean Winkler score
\end{description}
\endgroup

\vskip 0.2in
\bibliography{references}

\end{document}